\documentclass[letterpaper, 10 pt, conference]{ieeeconf}  

\IEEEoverridecommandlockouts                              
\usepackage{mathptmx} 
\usepackage{times}    

\usepackage{amsmath,amssymb}

\usepackage{graphicx}

\usepackage{booktabs}
\usepackage{array}
\usepackage{multirow}
\usepackage{xcolor}
\usepackage{colortbl}

\usepackage{stfloats} 

\usepackage{algorithm}
\usepackage{algpseudocode}

\usepackage{adjustbox}
\usepackage{tikz}
\usepackage[normalem]{ulem} 
\usepackage{changes}

\usepackage{cite}

\usepackage{cuted}
\usepackage{capt-of} 

\usepackage[breaklinks,colorlinks]{hyperref}

\title{\LARGE \bf
AgriGS-SLAM: Orchard Mapping Across Seasons\\ via Multi-View Gaussian Splatting SLAM
}

\author{
  Mirko Usuelli$^{1,\dagger}$, 
  David Rapado-Rincon$^{2}$, 
  Gert Kootstra$^{2}$, 
  Matteo Matteucci$^{1}$
  \thanks{$^{\dagger}$ Corresponding author; $^{1}$Mirko Usuelli and Matteo Matteucci are with the Dipartimento di Bioingegneria, Elettronica e Informazione, Politecnico di Milano, 20133 Milano, Italy
    {\tt\small \{mirko.usuelli, matteo.matteucci\}@polimi.it}}%
  \thanks{$^{2}$David Rapado-Rincon and Gert Kootstra are with Agricultural Biosystems Engineering, Wageningen University \& Research, 6708 PB Wageningen, The Netherlands
    {\tt\small \{david.rapadorincon, gert.kootstra\}@wur.nl}}%
}

\begin{document}

\maketitle

\begin{strip}
  \centering
  \vspace{-8.0em}  
  \includegraphics[width=\textwidth]{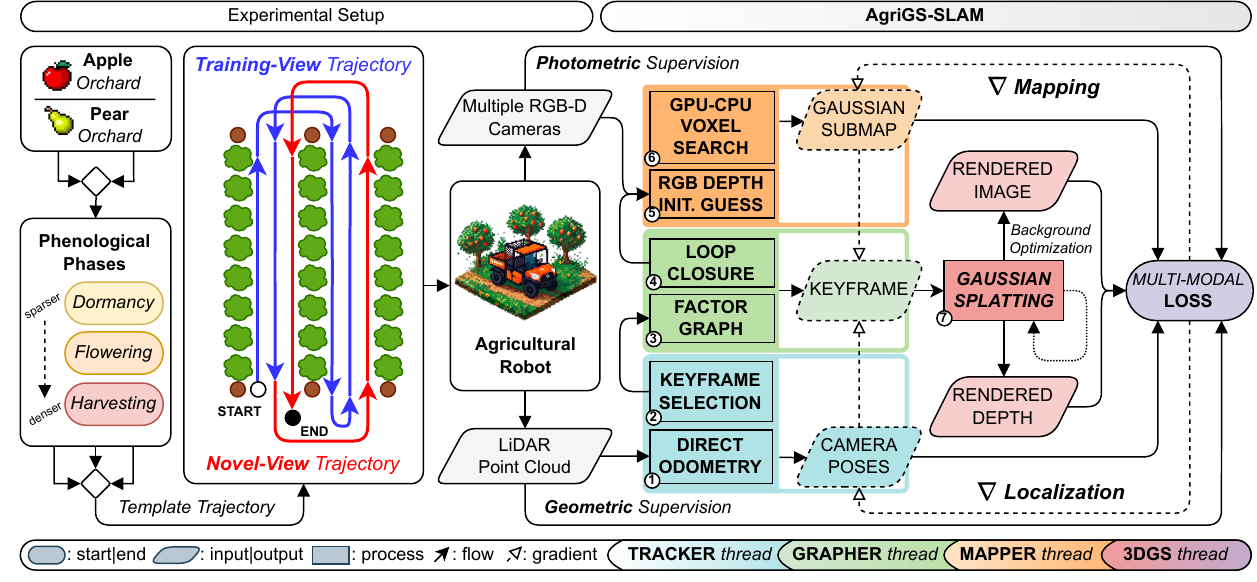}
  \vspace{-2.3em}
  \captionof{figure}{The agricultural robot follows a standardized trajectory (Training- and Novel-View) in both apple and pear orchards, across dormancy, flowering, and harvesting stages. LiDAR odometry keyframes update a loop-closing factor graph that triggers background 3DGS optimization, jointly refining Gaussian submaps and camera poses. Rendered depths and images are corrected via a multimodal geometric and photometric loss, enabling simultaneous gradient-based refinement for both localization and mapping.}\label{fig:cover}
\vspace{-1.3em}
\end{strip}

\begin{abstract}
Autonomous robots in orchards require real-time 3D scene understanding despite repetitive row geometry, seasonal appearance changes, and wind-driven foliage motion. We present AgriGS-SLAM, a Visual--LiDAR SLAM framework that couples direct LiDAR odometry and loop closures with multi-camera 3D Gaussian Splatting (3DGS) rendering. Batch rasterization across complementary viewpoints recovers orchard structure under occlusions, while a unified gradient-driven map lifecycle executed between keyframes preserves fine details and bounds memory. Pose refinement is guided by a probabilistic LiDAR-based depth consistency term, back-propagated through the camera projection to tighten geometry-appearance coupling. We deploy the system on a field platform in apple and pear orchards across dormancy, flowering, and harvesting, using a standardized trajectory protocol that evaluates both training-view and novel-view synthesis to reduce 3DGS overfitting in evaluation. Across seasons and sites, AgriGS-SLAM delivers sharper, more stable reconstructions and steadier trajectories than recent state-of-the-art 3DGS-SLAM baselines while maintaining real-time performance on-tractor. While demonstrated in orchard monitoring, the approach can be applied to other outdoor domains requiring robust multimodal perception.
\end{abstract}


\section{Introduction}
\label{sec:introduction}

The rapid growth of the global population is driving an urgent demand for increased food production, while labor shortages and strict sustainability regulations increase the need for autonomous farming technologies~\cite{kootstra2021selective}. In response, autonomous systems equipped with advanced sensing and decision-making capabilities are being developed to reduce dependence on seasonal labor and improve the efficient use of resources. Robotic automation in orchards presents particular challenges that necessitate accurate 3D reconstruction capabilities. Every tree develops differently and fruit trees undergo continuous transformations across the growing season: foliage density shifts dramatically, blossoms emerge and fade, and fruit develops over time. Temporal changes require methods that can capture and maintain accurate spatial representations. 

While offline Structure-from-Motion (SfM) approaches can generate high-quality reconstructions in post-processing, agricultural applications demand online and real-time capabilities for several critical reasons. First, autonomous agricultural robots require immediate spatial understanding for navigation and manipulation tasks: they cannot wait for offline processing to complete harvesting, spraying, or pruning operations. Second, farmers need immediate feedback during field operations to adjust interventions on the fly, such as modifying spray patterns based on current canopy density or identifying areas requiring instant attention. Third, the concept of digital twins in agriculture envisions continuous synchronization between the physical world and its digital representations, enabling interactive monitoring. To simultaneously localize the robot and build maps in real-time, these requirements necessitate SLAM (Simultaneous Localization and Mapping).

There is a significant research gap in the development of SLAM systems suitable for large-scale agricultural environments, such as orchards. The repetitive geometry of tree rows introduces systematic ambiguities in short- and long-term associations, where the scarcity of distinctive landmarks often leads to incorrect data associations. Existing SLAM approaches remain limited in this context: vision-only methods fail under repetitive crop patterns and vegetation motion~\cite{Shu_2021_WACV}, while LiDAR-only systems are hampered by geometric sparsity compared to the structured surfaces of man-made environments~\cite{Chen_2020_RAL}. 


Recent advances in neural rendering have introduced promising paradigms for dense 3D reconstruction. Neural Radiance Fields (NeRF)~\cite{mildenhall2020nerf} encode scenes as neural functions but suffer from expensive training and poor scalability. In contrast, 3D Gaussian Splatting (3DGS)~\cite{kerbl20233d} offers explicit point-based representations with discrete Gaussian primitives, called splats, enabling localized updates and efficient rasterization, properties well-suited for the incremental nature of SLAM in large scenes. However, most NeRF and 3DGS-based SLAM methods remain restricted to controlled room-scale indoor scenes~\cite{sh-ch18-spatial-ai}. 

\begin{figure}[t]
\vspace{1.0em}
\centering
\includegraphics[width=0.48\textwidth]{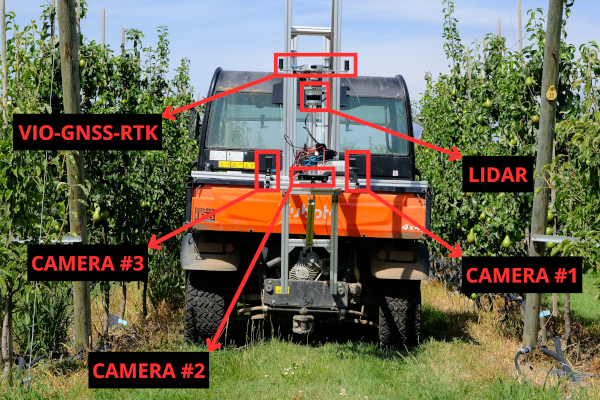}
\caption{Agricultural platform equipped with three cameras (one horizontal and two vertical), a 32-beam LiDAR, and a VIO–GNSS–RTK system for ground truth acquisition.}\label{fig:setup}
\vspace{-2.0em}
\end{figure}

Building on recent advances in SLAM and 3D reconstruction, we propose a Visual--LiDAR SLAM framework (Fig.~\ref{fig:cover}) tailored to orchards. A key contribution is the multi-camera setup, which provides complementary perspectives around the vehicle: lateral views improve canopy coverage under occlusions, while the inline view supports traversability perception along narrow rows (Fig.~\ref{fig:setup}). On the geometric side, direct LiDAR odometry delivers robust tracking, remaining largely unaffected by wind-induced foliage motion or illumination changes that can disrupt vision methods, since LiDAR operates on the point cloud structure as a whole. The two modalities are fused to both improve 3DGS rendering and refine camera poses, with loop closures detected and refined through scan-matching to correct drift during extended row traversals and tight inter-row maneuvers.

The main contributions of this work are:
\begin{itemize}
    \item \textbf{Real-Time Visual--LiDAR 3DGS-SLAM (\ref{sec:proposed_approach}):}\\
    Our method combines LiDAR odometry, loop closure, multimodal optimization, and memory-aware incremental 3D Gaussian Splatting, providing real-time perception, 3D reconstruction, and trajectory estimation for agricultural robots in orchards.

\item \textbf{Cross-Seasonal Applicability Benchmark (\ref{sec:experiments},~\ref{sec:results}):}  
    \\We compare against diverse 3DGS-SLAM state-of-the-art pipelines (Photo-SLAM~\cite{Huang_2024_CVPR}, Splat-SLAM~\cite{sandstrom2024splatslam}, OpenGS-SLAM~\cite{yu2025rgb}, PINGS~\cite{pan2025pings}). Our pipeline evaluates both training- and novel-view trajectories, overcoming prior methods that only assess training-view rendering. This enables a fair joint evaluation of reconstruction and localization, highlighting the generalization and performance gains of our approach across seasons.


    \item \textbf{$\nabla$Mapping \& $\nabla$Localization (\ref{subsec:ablation} \S\ref{subsec:ablation:localization}, \S\ref{subsec:ablation:mapping}):} \\A unified, gradient-driven 3DGS-SLAM that (i) updates the map between keyframes via image-gradient densification, opacity/scale-aware pruning, and scheduled opacity resets under per-batch budgets; and (ii) refines poses using a probabilistic LiDAR-guided depth-consistency loss, defined as a Kullback–Leibler (KL) divergence between LiDAR and rendered depth, enabling sensor fusion with cameras and back-propagation through the projection model.

    \item \textbf{Multi-Camera 3DGS-SLAM (\ref{subsec:ablation} \S\ref{subsec:ablation:multicamera}):} \\To the best of the authors' knowledge, the first real-time 3DGS-SLAM framework supporting multi-camera setups in outdoor environments, enabled by a batch rasterization strategy. Multiple cameras are leveraged to overcome occlusions and limited viewpoints in orchard.

\end{itemize}
The entire implementation of this work is open source and
available on GitHub at:~\url{https://github.com/AIRLab-POLIMI/agri-gs-slam}.

\begin{table*}[t]
\vspace{1.0em}
\centering
\caption{Multi-view image comparison grouped by orchard type, phenological stage and rendering modality.}
\label{tab:slam_multiview_grouped_ocm}
\renewcommand{\arraystretch}{1.2}
\setlength{\tabcolsep}{2pt}

\begin{tabular}{@{}c*{4}{c}@{}}
\toprule
& \multicolumn{2}{c}{\textbf{Apple Orchard}} & \multicolumn{2}{c}{\textbf{Pear Orchard}} \\
\cmidrule(lr){2-3} \cmidrule(lr){4-5}
& GT / \textbf{Training}-View Rendering & GT / \textbf{Novel}-View Rendering & GT / \textbf{Training}-View Rendering & GT / \textbf{Novel}-View Rendering \\
\midrule

\rotatebox{90}{\textbf{Dormancy}} &
\includegraphics[width=0.24\textwidth]{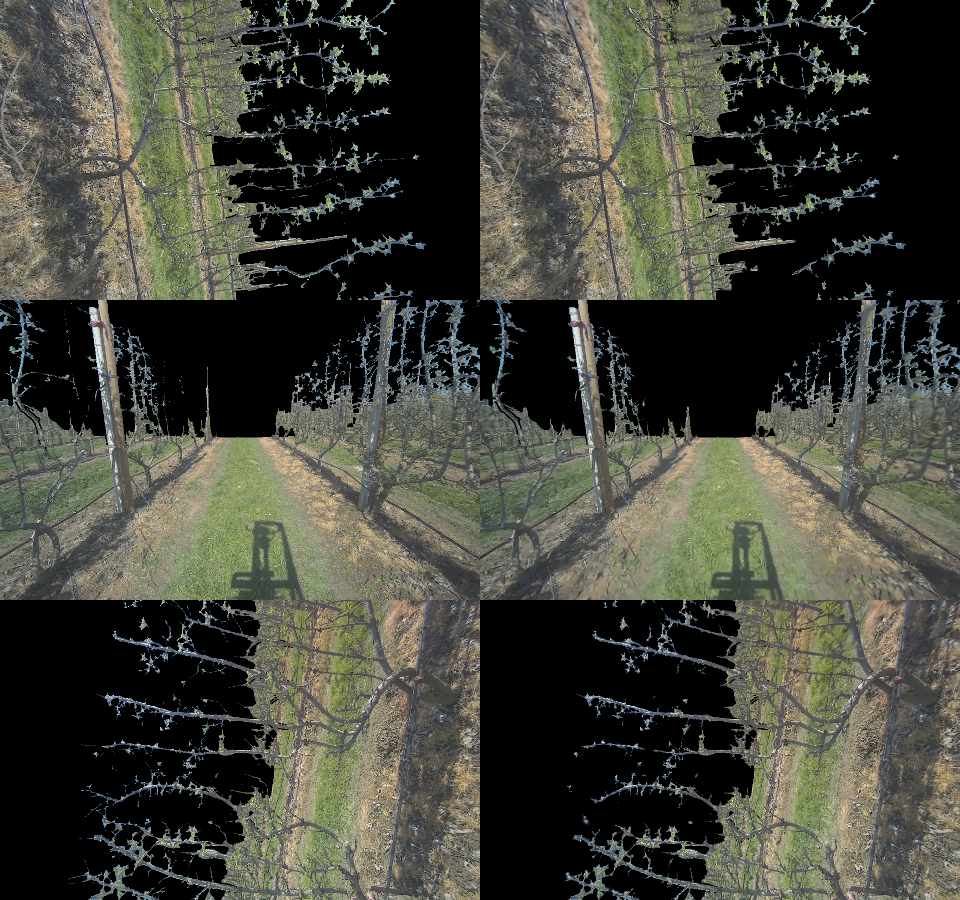} &
\includegraphics[width=0.24\textwidth]{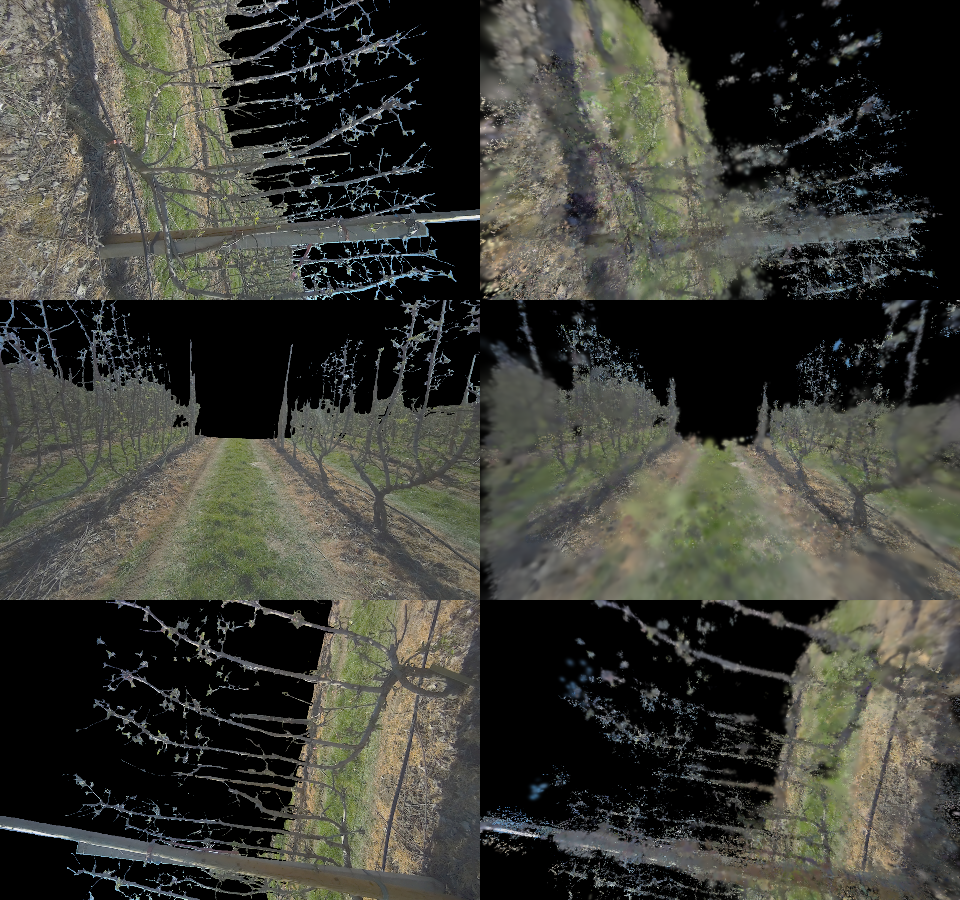} &
\includegraphics[width=0.24\textwidth]{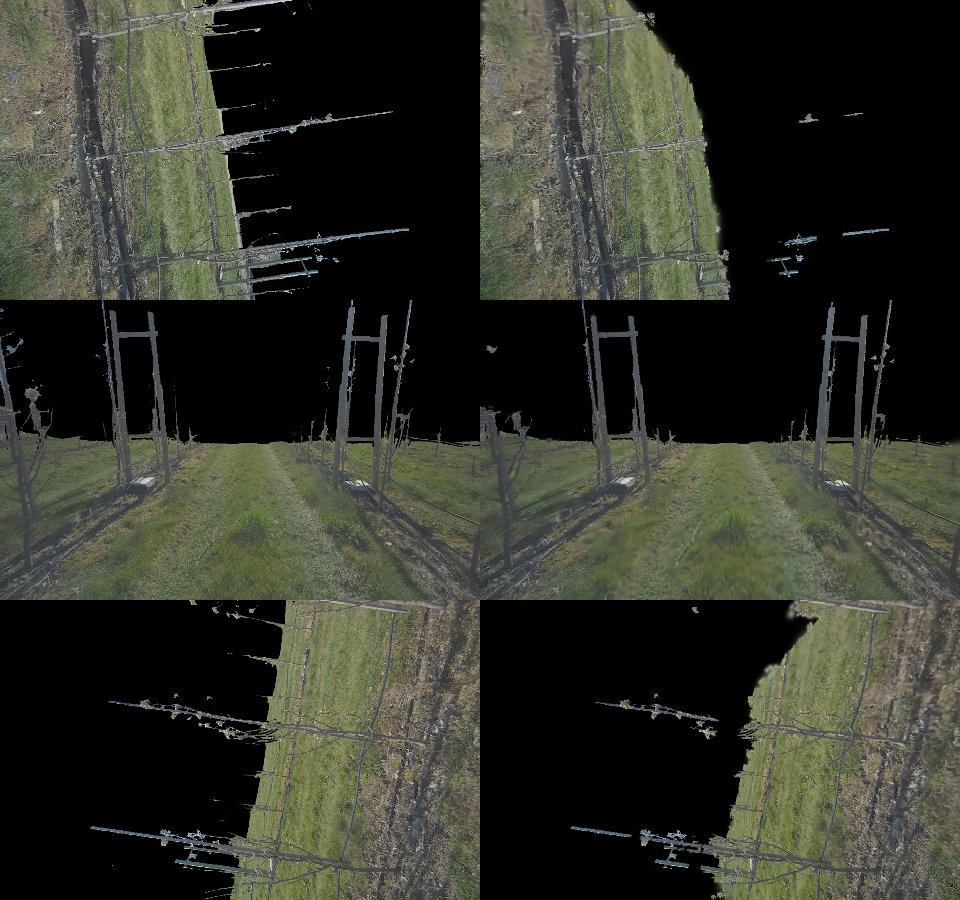} &
\includegraphics[width=0.24\textwidth]{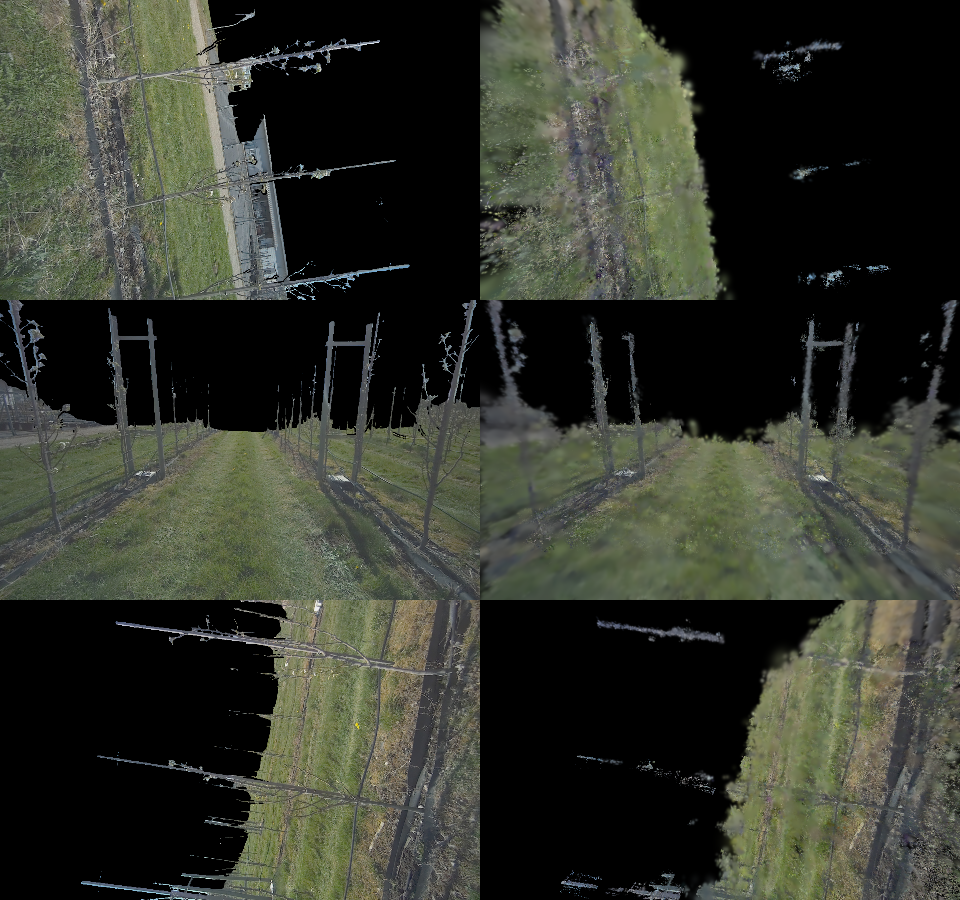} \\

\rotatebox{90}{\textbf{Flowering}} &
\includegraphics[width=0.24\textwidth]{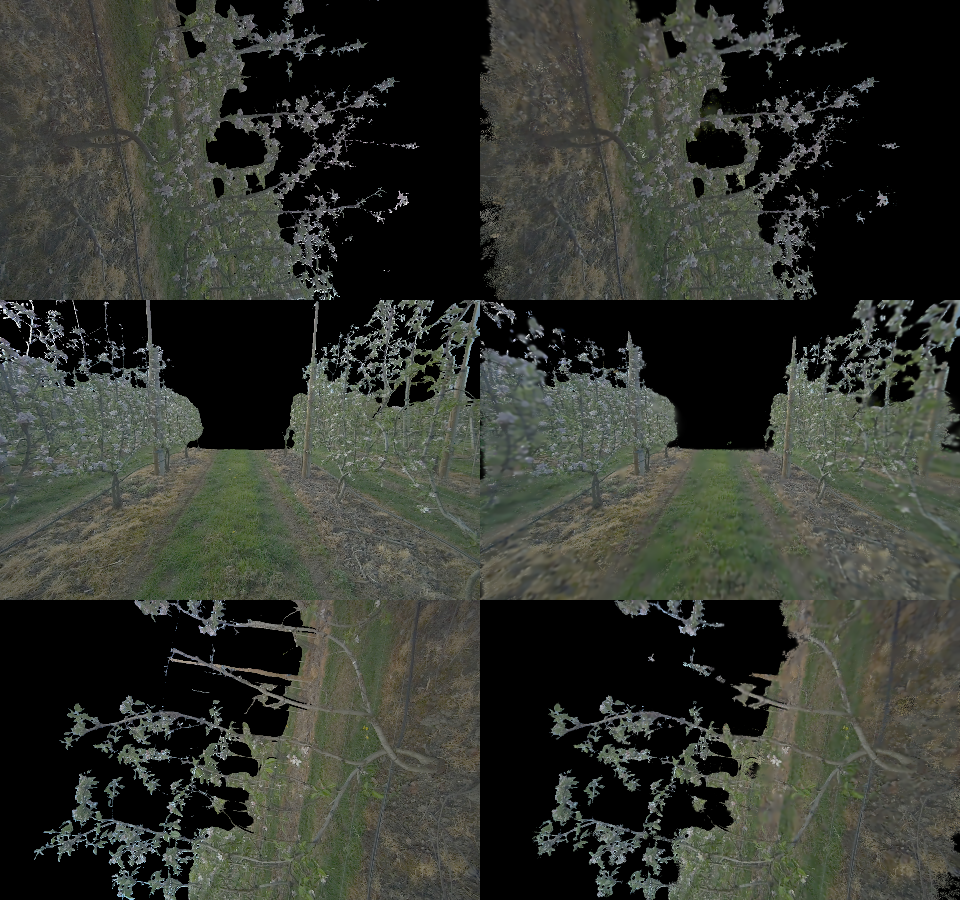} &
\includegraphics[width=0.24\textwidth]{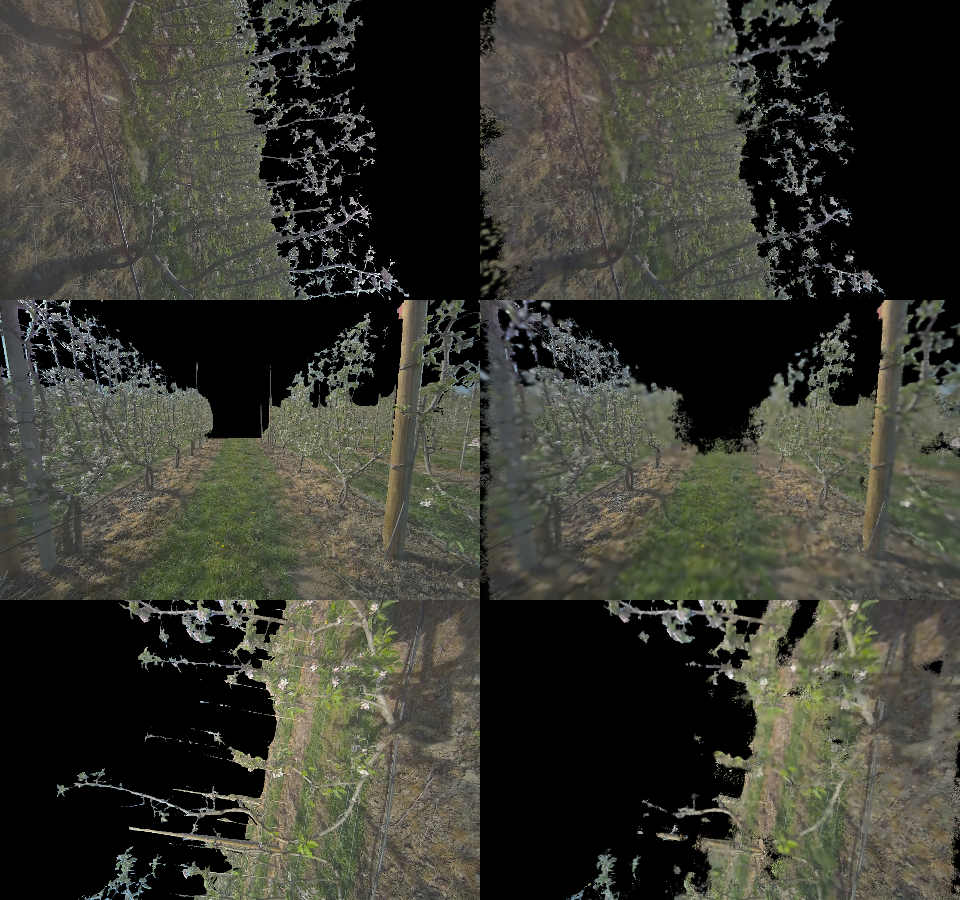} &
\includegraphics[width=0.24\textwidth]{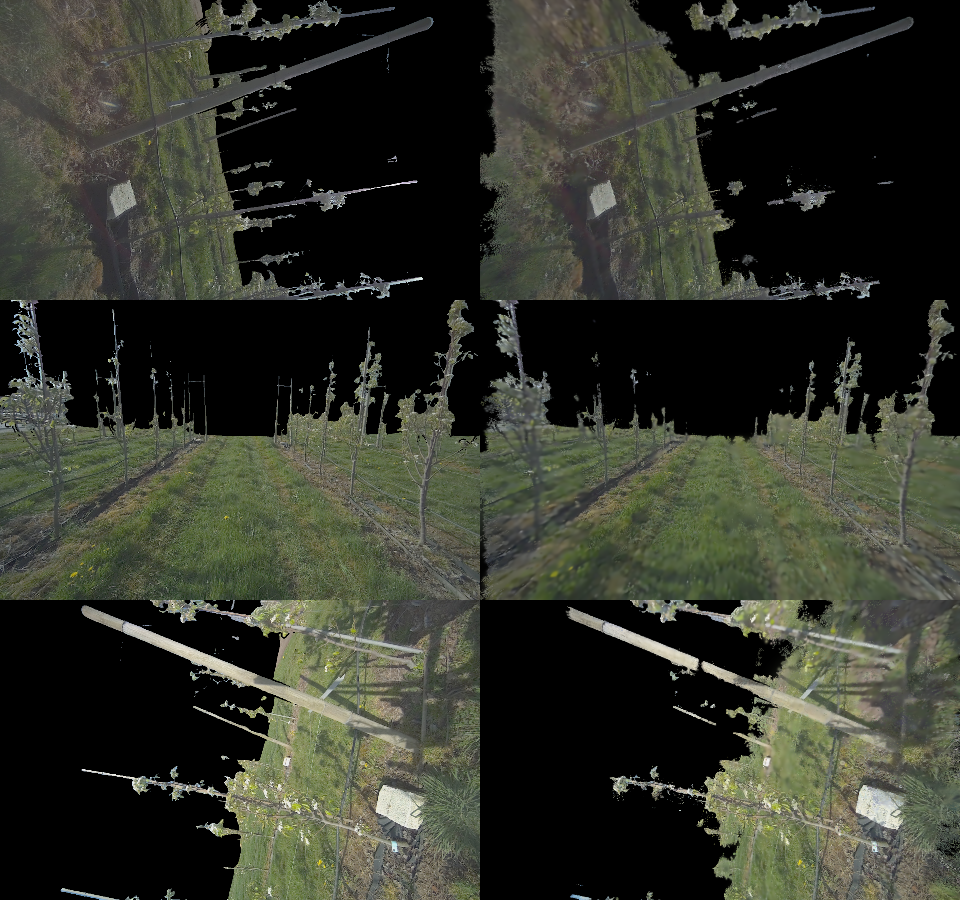} &
\includegraphics[width=0.24\textwidth]{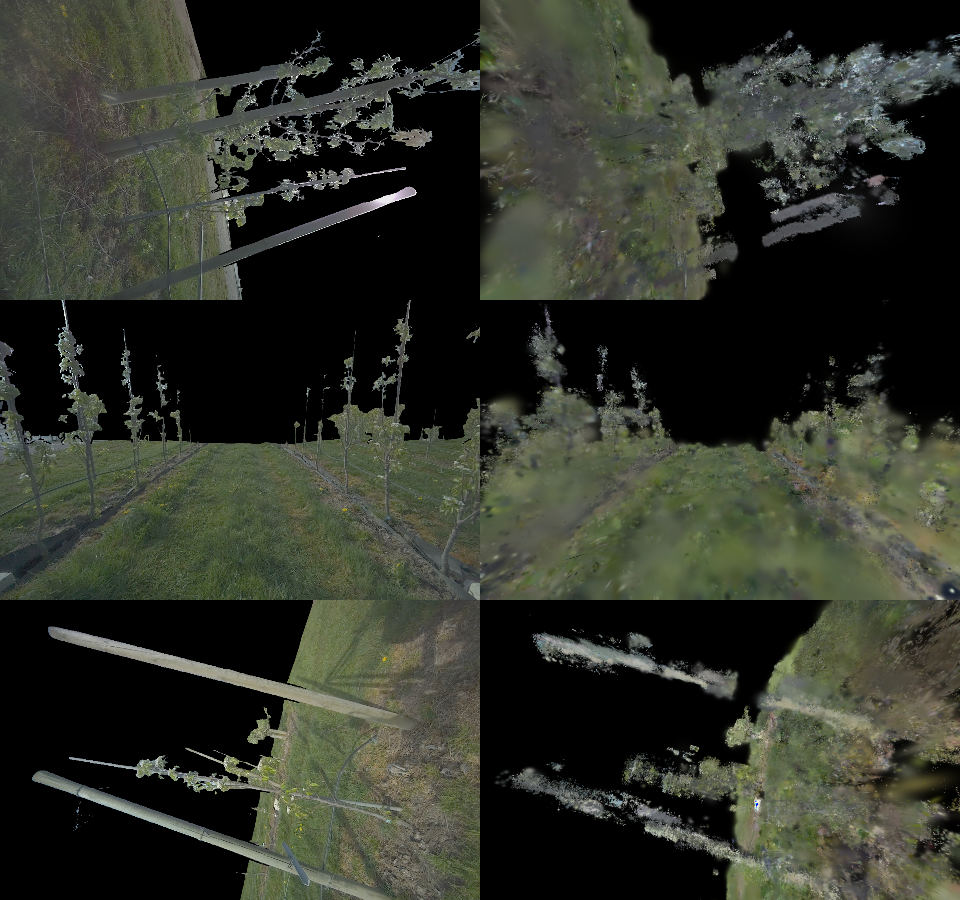} \\

\rotatebox{90}{\textbf{Harvesting}} &
\includegraphics[width=0.24\textwidth]{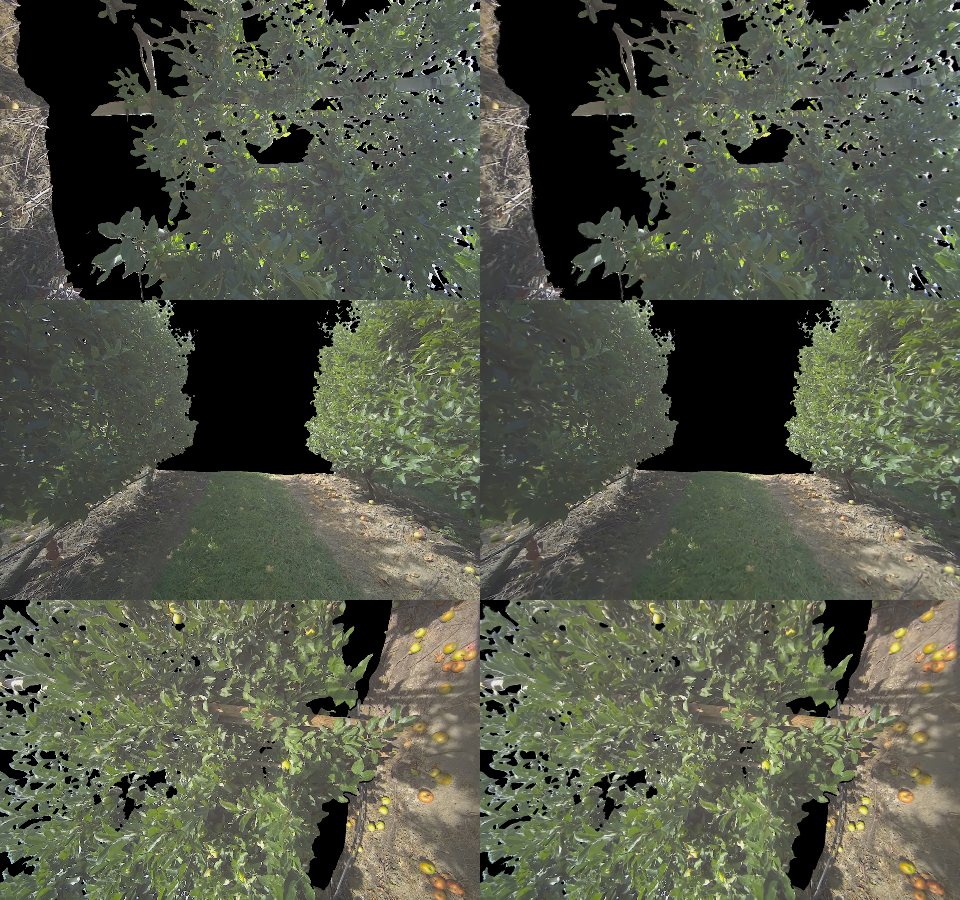} &
\includegraphics[width=0.24\textwidth]{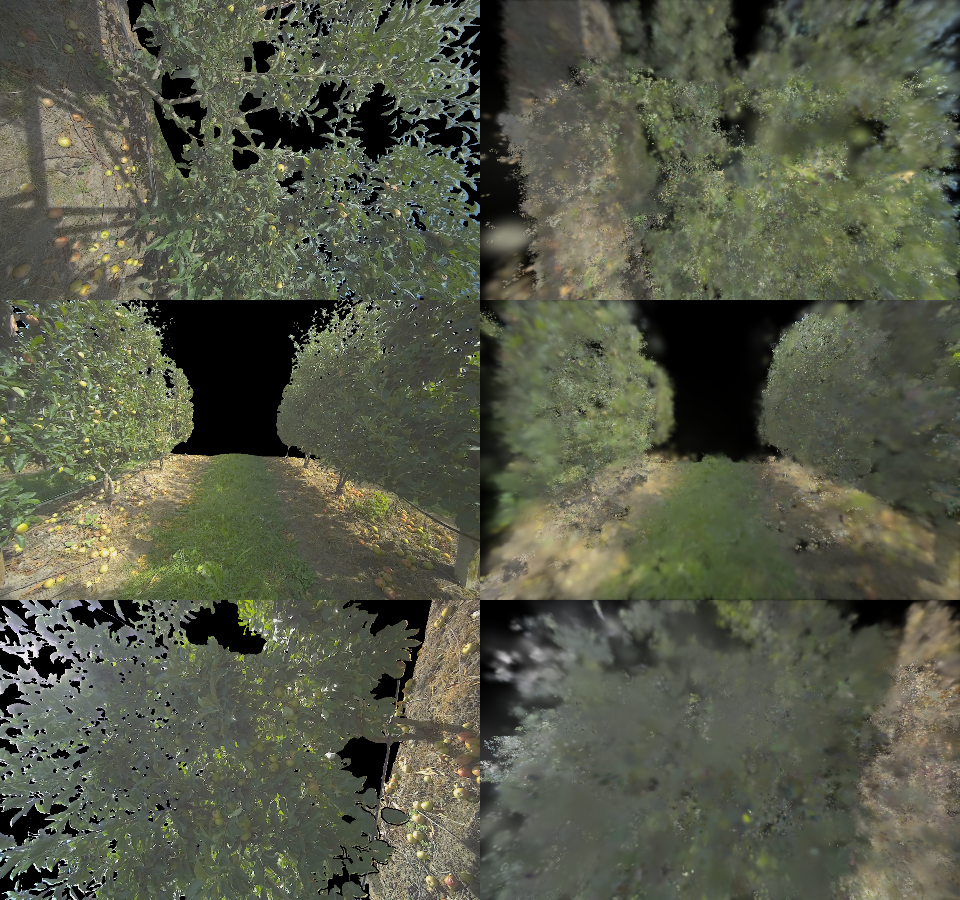} &
\includegraphics[width=0.24\textwidth]{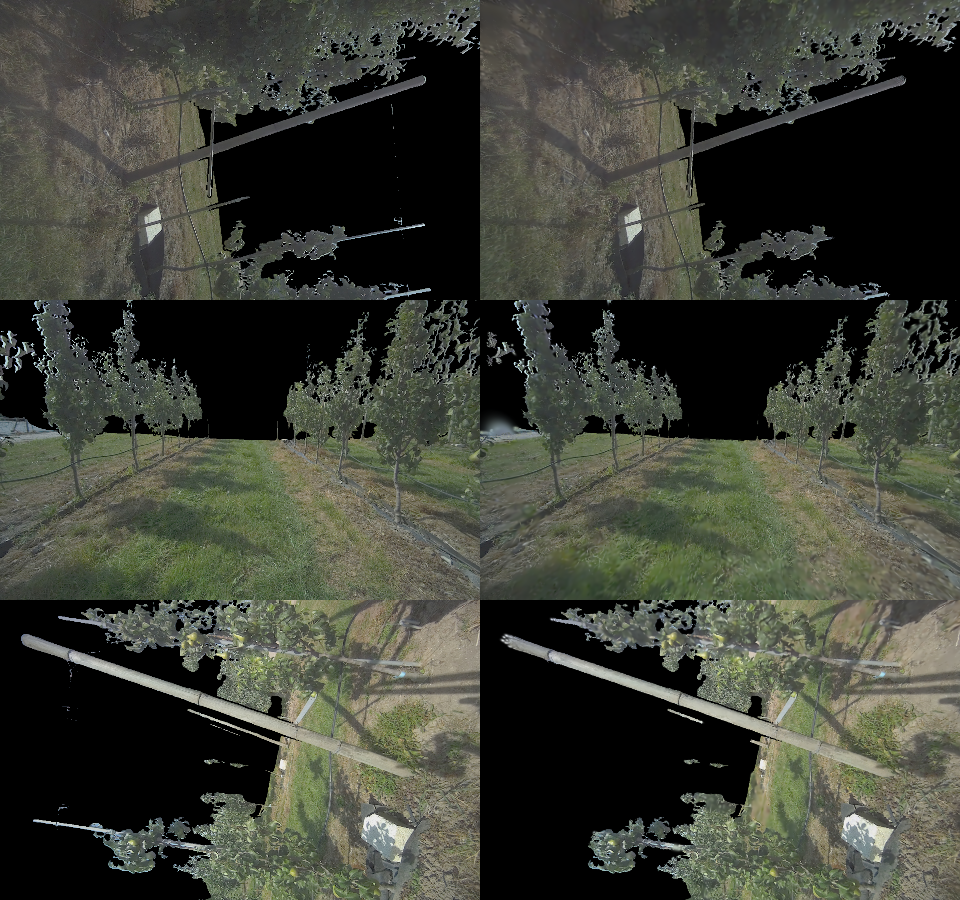} &
\includegraphics[width=0.24\textwidth]{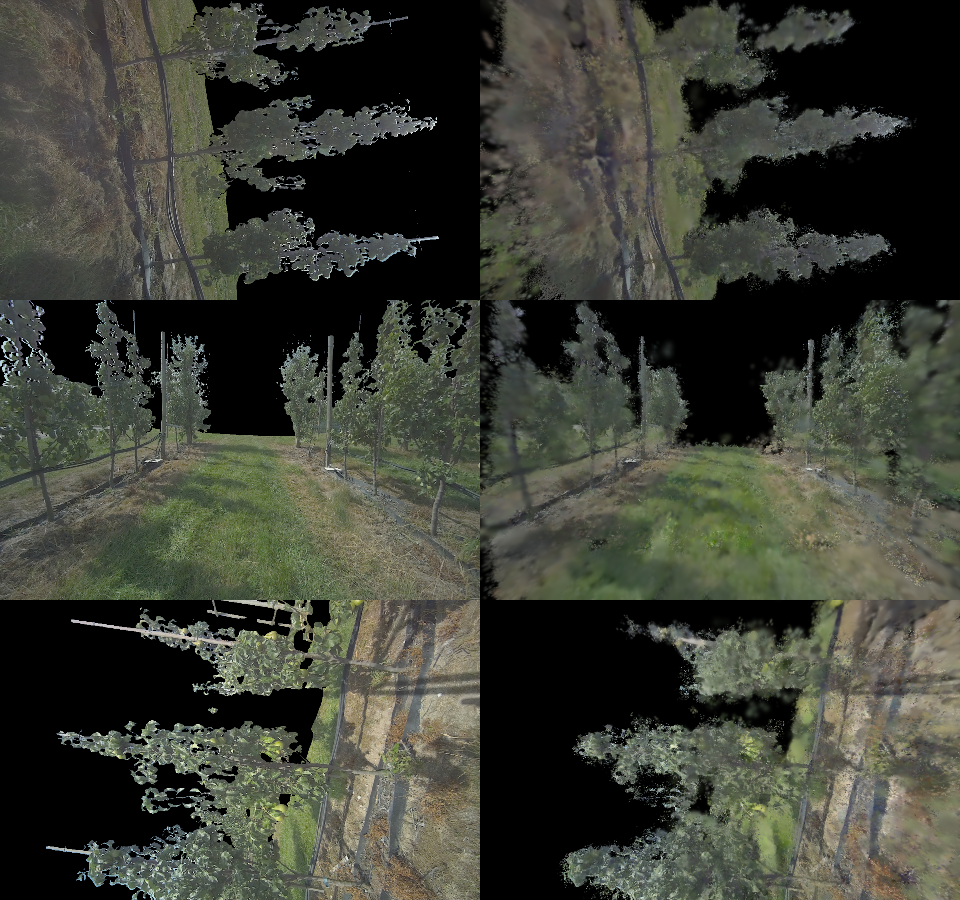} \\

\bottomrule
\end{tabular}
\renewcommand{\arraystretch}{1.0}
\setlength{\tabcolsep}{6pt}
\vspace{-2.0em}
\end{table*}

\vspace{-0.3em}
\section{Related Work}
\label{sec:related_work}

Recent advances in neural rendering have enriched the traditional SLAM pipeline, coupling accurate localization with photorealistic 3D reconstruction. Early implicit methods based on neural radiance fields, such as NeRF-SLAM~\cite{nerfslam2022}, encoded geometry and appearance directly in network weights, achieving dense reconstructions with strong photometric consistency. However, these approaches proved computationally prohibitive for real-time robotics applications, scaling poorly to large environments~\cite{sh-ch18-spatial-ai}.

To overcome these limitations, 3DGS~\cite{kerbl20233d} emerged as an explicit alternative. Originally developed as an extension of classical photogrammetry, 3DGS reformulates multi-view reconstruction into a differentiable scene representation. By modeling environments as collections of Gaussian primitives, it preserves the flexibility of optimization while allowing localized updates. This formulation has since provided the foundation for adapting 3DGS to SLAM, enabling practical real-time systems that couple reconstruction and localization.

Adapting 3DGS-SLAM to outdoor settings introduced substantial challenges, including variable illumination, dynamic elements, and sparse or absent depth measurements. Two primary strategies have emerged to address these constraints: sensor-fusion methods, such as LiV-GS~\cite{liVGS2024}, which integrate point clouds into 3DGS with geometric and normal consistency under inertial correction, and classical visual pipelines, such as Splat-SLAM~\cite{sandstrom2024splatslam}, which rely solely on RGB inputs to infer geometry via multi-view optimization or learned depth priors without dense depth supervision.

Recent systems exhibit considerable architectural diversity, reflecting different integration strategies and algorithmic philosophies. Photo-SLAM~\cite{Huang_2024_CVPR} extends the classical ORB-SLAM3~\cite{campos2021orb} framework by augmenting feature-based tracking with Gaussian rendering for enhanced photometric refinement. Splat-SLAM~\cite{sandstrom2024splatslam} integrates 3DGS directly into the dense bundle adjustment pipeline of DROID-SLAM~\cite{teed2021droid}. OpenGS-SLAM~\cite{yu2025rgb} combines transformer-based depth estimation with RANSAC-PnP for robust pose estimation in challenging conditions, while PINGS~\cite{pan2025pings} fuses LiDAR and SDF implicit points into splats to achieve joint geometric and photometric alignment. This heterogeneous landscape illustrates the rapid maturation of 3DGS-SLAM from initial proof-of-concept systems to a diverse family of pipelines spanning classical and multimodal paradigms.

Within agricultural robotics, 3D reconstruction has primarily served offline tasks such as phenotyping, yield estimation, and crop perception~\cite{fruitnerf2023,grun2025towards, 11163998, agrinerf2024}. While NeRF-based approaches have been explored for agricultural scenes, the distinctive challenges of orchard environments remain largely unaddressed in online 3DGS-SLAM research. These challenges include highly repetitive row structures, frequent occlusions from dense foliage, and pronounced seasonal appearance variations that violate typical photometric consistency assumptions. 

\section{Proposed Approach}
\label{sec:proposed_approach}


\begin{table*}[t]
\vspace{1.0em}
\centering
\small
\renewcommand{\arraystretch}{1.2}
\setlength{\tabcolsep}{4pt}
\begin{adjustbox}{max width=\textwidth}
\begin{tabular}{ll
  *{7}{c}  
  *{7}{c}  
  *{7}{c}  
}
\toprule
& & 
\multicolumn{7}{c}{\textbf{Dormancy}} & 
\multicolumn{7}{c}{\textbf{Flowering}} & 
\multicolumn{7}{c}{\textbf{Harvesting}} \\
\cmidrule(lr){3-9} \cmidrule(lr){10-16} \cmidrule(lr){17-23}
& & 
\multicolumn{3}{c}{\textit{\textbf{Training}-View}} & \multicolumn{3}{c}{\textit{\textbf{Novel}-View}} &  
& \multicolumn{3}{c}{\textit{\textbf{Training}-View}} & \multicolumn{3}{c}{\textit{\textbf{Novel}-View}} &  
& \multicolumn{3}{c}{\textit{\textbf{Training}-View}} & \multicolumn{3}{c}{\textit{\textbf{Novel}-View}} &  
\\
\cmidrule(lr){3-5} \cmidrule(lr){6-8}
\cmidrule(lr){10-12} \cmidrule(lr){13-15}
\cmidrule(lr){17-19} \cmidrule(lr){20-22}
 & Method
& PSNR $\uparrow$ & SSIM $\uparrow$ & LPIPS $\downarrow$ & PSNR $\uparrow$ & SSIM $\uparrow$ & LPIPS $\downarrow$ & ATE (m) $\downarrow$
& PSNR $\uparrow$ & SSIM $\uparrow$ & LPIPS $\downarrow$ & PSNR $\uparrow$ & SSIM $\uparrow$ & LPIPS $\downarrow$ & ATE (m) $\downarrow$
& PSNR $\uparrow$ & SSIM $\uparrow$ & LPIPS $\downarrow$ & PSNR $\uparrow$ & SSIM $\uparrow$ & LPIPS $\downarrow$ & ATE (m) $\downarrow$
\\
\midrule
\multirow{8}{*}{\rotatebox{90}{\textbf{Apple Orchard}}}
& Photo-SLAM~\cite{Huang_2024_CVPR}   &  8.1567  &  0.2273  &  0.7058  & 8.3705  &  0.2137  &  0.7466 & 18.9545  &  8.9779 &  0.2990 &  0.7077  &  9.4701  & 0.2604 &  0.7424  & 15.1569  &  9.2807  &  0.3442  & 0.6823 & 9.7843 & 0.3211 &  0.7278  &  19.0975   \\
& Splat-SLAM~\cite{sandstrom2024splatslam}   &  19.6937  &  0.1761  &  0.9562  & 11.0302  &  0.1575  &  0.9997 & 5.2648  &  19.7930 &  0.1893 &  0.9879  &  12.3704  & 0.1255  &  0.9952  & 28.1355   &  19.7298  &  0.2730  &  0.9561 & 14.5519  & 0.2567  &  0.9844  &  21.8882   \\
& PINGS~\cite{pan2025pings}    &  10.7060  &  0.3030  &  0.6343  & 10.1925  &  0.2448  &  0.6908 & 171.4970  &  6.2423 &  0.1822 &  0.6839  &  6.8555  & 0.1528  &  0.7208  & 132.9355   &  5.3410  &  0.2325  & 0.6865 & 4.3199 & 0.2064   &  0.7112  &  89.4429   \\

& OpenGS-SLAM~\cite{yu2025rgb}    &  12.9748  &  0.3486  &  0.7627  & 14.5637  &  \cellcolor{red!30}\textbf{0.4231}  &  0.7772 & 20.6953  &  14.3540 &  0.3732 &  0.7746  & 14.2038  & 0.3553 & 0.7995  & 20.9450 &  14.3598  &  0.4346  & 0.7033 & 14.0214 & 0.3984 &  0.7403 &  20.2172 \\

& DLO+3DGS~\cite{9681177, kerbl20233d}    &  25.0465  &  0.7157  &  0.3854  & \cellcolor{red!30}\textbf{15.8479}  &  0.3781  &  0.8092 & \cellcolor{yellow!30}0.5756  &  23.2178 &  0.6680 &  0.4989  &  \cellcolor{orange!30}\underline{16.5182}  & \cellcolor{yellow!30}0.3664   &  0.7929  & 0.4084   &  31.6495  &  0.8595  &  0.2141 & \cellcolor{red!30}\textbf{18.0344}   & \cellcolor{orange!30}\underline{0.4541}   &  0.7998  &  \cellcolor{red!30}\textbf{0.6578}   \\

& \textbf{AgriGS-SLAM} &  \cellcolor{orange!30}\underline{29.8954}  &  \cellcolor{yellow!30}0.8959  &  \cellcolor{orange!30}\underline{0.1316}  & \cellcolor{orange!30}\underline{15.7438}  &   \cellcolor{yellow!30}{0.3890}  &  \cellcolor{orange!30}\underline{0.5915} & \cellcolor{orange!30}\underline{0.5192}  &  \cellcolor{yellow!30}{25.5346} &  \cellcolor{orange!30}\underline{0.8239} &  \cellcolor{yellow!30}{0.2319}  &  \cellcolor{red!30}\textbf{16.7627}  & \cellcolor{red!30}\textbf{0.3878}   &  \cellcolor{red!30}\textbf{0.5410}  & \cellcolor{orange!30}\underline{0.3543}   &  \cellcolor{yellow!30}31.8057  &  \cellcolor{yellow!30}0.8755  & \cellcolor{yellow!30}0.1599 & \cellcolor{orange!30}\underline{17.8304}  & \cellcolor{red!30}\textbf{0.4605}   &  \cellcolor{red!30}\textbf{0.6237}  &  0.7370   \\

& AgriGS w/2 cameras &  \cellcolor{yellow!30}26.2831  &  \cellcolor{orange!30}\underline{0.9129}  &  \cellcolor{yellow!30}0.1341  & 14.4759  &  0.3739  &  \cellcolor{yellow!30}0.6018 & \cellcolor{orange!30}\underline{0.5192}  &  \cellcolor{orange!30}\underline{26.8023} &  \cellcolor{yellow!30}0.9006 &  \cellcolor{orange!30}\underline{0.1571}  &  \cellcolor{yellow!30}15.6297  & \cellcolor{orange!30}\underline{0.3808}   &  \cellcolor{orange!30}\underline{0.5642}  & \cellcolor{yellow!30}0.3561   &  \cellcolor{orange!30}\underline{33.5170}  &  \cellcolor{orange!30}\underline{0.9157}  &    \cellcolor{orange!30}\underline{0.1127} & \cellcolor{yellow!30}15.7548   & \cellcolor{yellow!30}0.4077   &  \cellcolor{orange!30}\underline{0.6646}  &  \cellcolor{yellow!30}0.7348   \\

& AgriGS w/3 cameras &  \cellcolor{red!30}\textbf{30.0287}  &   \cellcolor{red!30}\textbf{0.9249}  &  \cellcolor{red!30}\textbf{0.1253}  & \cellcolor{yellow!30}14.6592  &  \cellcolor{orange!30}\underline{0.4153}  &  \cellcolor{red!30}\textbf{0.5500} & \cellcolor{red!30}\textbf{0.5104}  &  \cellcolor{red!30}\textbf{28.6355} &  \cellcolor{red!30}\textbf{0.9137} &  \cellcolor{red!30}\textbf{0.1367}  &  14.7663  & 0.3411   &  \cellcolor{yellow!30}0.5667  & \cellcolor{red!30}\textbf{0.3410}   &  \cellcolor{red!30}\textbf{35.1030}  &  \cellcolor{red!30}\textbf{0.9307}  & \cellcolor{red!30}\textbf{0.1064} & 15.1278   & 0.3727   &  \cellcolor{yellow!30}0.6705  &  \cellcolor{orange!30}\underline{0.7281}   \\
\midrule
\multirow{8}{*}{\rotatebox{90}{\textbf{Pear Orchard}}}
& Photo-SLAM~\cite{Huang_2024_CVPR}   &  7.9990  &  0.2739  &  0.6536  & 7.9589  &  0.2678  &  0.6382 & 16.2391  &  7.8072 &  0.2801 &  0.6565  &  8.2488  & 0.3417   &  0.6619  & 23.7226   &  8.3235  &  0.2839  &  0.6513 & 8.1888 & 0.2462 &  0.6713  &  18.9968   \\
& Splat-SLAM~\cite{sandstrom2024splatslam}   &  21.5711*  &  0.2170*  &  0.9078*  & 11.1528*  &  0.1626*  &  0.9448* & 6.6386*  &  23.5328 &  0.2829 &  0.6885  &  13.9598  & 0.2822 &  0.8749  & 20.8523 &  18.4055  &  0.2114  & 0.9446 & 13.5946 & 0.1635 &  0.9468  &  18.2869   \\
& PINGS~\cite{pan2025pings}    &  9.2289  &  0.3402  &  0.6027  & 9.7279  &  0.3449  &  0.5964 & 157.2260  &  13.0218 &  0.4558 &  0.5374  &  12.4485  & 0.4538  &  0.5615  & 87.3478 &  11.7895  &  0.3322  & 0.6089 & 10.8935 & 0.2549 &  0.6858  &  97.7556   \\
& OpenGS-SLAM~\cite{yu2025rgb}    &  12.8747 &  0.3920 & 0.6911  & 15.2439  & 0.5080  & 0.6410 & 21.0662  &  14.3015 &  0.4604 &  0.6722  &  16.1294  & \cellcolor{red!30}\textbf{0.5783}  &  0.6608  & 10.9321   &  15.3887  &  0.4354  & 0.6580 & 15.4138   & \cellcolor{yellow!30}0.4125   &  0.7008  &  20.8209  \\

& DLO+3DGS~\cite{9681177, kerbl20233d}    &  \cellcolor{orange!30}\underline{18.5079}  &  \cellcolor{yellow!30}0.6724  &  0.5774  & \cellcolor{red!30}\textbf{18.6592}  &  \cellcolor{orange!30}\underline{0.5243}  &  0.6613 & 0.4402  &  \cellcolor{yellow!30}30.5728 &  0.8867 &  0.2095  &  \cellcolor{red!30}\textbf{19.0106}  & 0.5111 &  0.6550  & 0.4478   &  33.1395  &  0.8498  &  0.2607 & \cellcolor{red!30}\textbf{17.5047}   & 0.3615   &  0.7678  &  0.7551   \\

& \textbf{AgriGS-SLAM} &  \cellcolor{yellow!30}18.3281  &  \cellcolor{orange!30}\underline{0.6728}  &  \cellcolor{yellow!30}0.5648  & \cellcolor{orange!30}\underline{17.7175}  &  \cellcolor{red!30}\textbf{0.5539}  &  \cellcolor{orange!30}\underline{0.5244} & \cellcolor{red!30}\textbf{0.4264}  &  \cellcolor{orange!30}\underline{33.1657} &  \cellcolor{orange!30}\underline{0.9206} &  \cellcolor{yellow!30}0.1339  &  \cellcolor{orange!30}\underline{18.8224}  & \cellcolor{orange!30}\underline{0.5635}   &  \cellcolor{yellow!30}0.5289  & \cellcolor{yellow!30}0.4398   &  \cellcolor{red!30}\textbf{35.0662}  &  \cellcolor{yellow!30}0.9038  & \cellcolor{yellow!30}0.1531 & \cellcolor{orange!30}\underline{17.4914}   & 0.4006   &  \cellcolor{orange!30}\underline{0.5920}  &  \cellcolor{yellow!30}0.7201   \\

& AgriGS w/2 cameras &  \cellcolor{red!30}\textbf{19.3567}  &  \cellcolor{red!30}\textbf{0.7223}  &  \cellcolor{red!30}\textbf{0.4141}  & 15.9485  &  \cellcolor{yellow!30}0.5111  &  \cellcolor{red!30}\textbf{0.5023} & \cellcolor{yellow!30}0.4313  &  29.6681 &  \cellcolor{yellow!30}0.9176 &  \cellcolor{orange!30}\underline{0.1252}  &  \cellcolor{yellow!30}17.1290  & 0.5393   &  \cellcolor{orange!30}\underline{0.5056}  & \cellcolor{orange!30}\underline{0.4321}   &  \cellcolor{yellow!30}34.9862  &  \cellcolor{red!30}\textbf{0.9210}  &    \cellcolor{red!30}\textbf{0.1152} & \cellcolor{yellow!30}17.2648   & \cellcolor{orange!30}\underline{0.4655}   &  \cellcolor{yellow!30}0.5937  &  \cellcolor{orange!30}\underline{0.7150}   \\

& AgriGS w/3 cameras &  16.8865  &  0.5625  &  \cellcolor{orange!30}\underline{0.4830}  & \cellcolor{yellow!30}16.1695  &  0.4271  &  \cellcolor{yellow!30}0.5631 & \cellcolor{orange!30}\underline{0.4303}  &  \cellcolor{red!30}\textbf{34.3933} &  \cellcolor{red!30}\textbf{0.9300} &  \cellcolor{red!30}\textbf{0.1019}  &  16.7364  & \cellcolor{yellow!30}{0.5559}   &  \cellcolor{red!30}\textbf{0.4864}  & \cellcolor{red!30}\textbf{0.4305}   &  \cellcolor{orange!30}\underline{34.3865}  &  \cellcolor{orange!30}\underline{0.9146}  &    \cellcolor{orange!30}\underline{0.1238} & 16.2442   & \cellcolor{red!30}\textbf{0.4664}   &  \cellcolor{red!30}\textbf{0.5566}  &  \cellcolor{red!30}\textbf{0.7120}   \\
\bottomrule
\end{tabular}
\end{adjustbox}
\caption{Comparison of apple and pear orchards across phenological phases. Training/validation metrics (PSNR, SSIM, LPIPS) and Absolute Trajectory Error (ATE, m) are shown. Arrows mark whether higher ($\uparrow$) or lower ($\downarrow$) is better. Best, second, and third results are marked in \textbf{bold} (red), \underline{underline} (orange), and yellow. Asterisk ($*$) denotes partial trajectories due to divergence.}\label{tab:quantitative}
\vspace{-1.5em}
\end{table*}

\subsection{Simultaneous Localization and Mapping}

\subsubsection{SLAM Frontend}
\label{sec:slam_frontend}
We use Direct LiDAR Odometry (DLO)~\cite{9681177} as the scan-to-submap odometry module, since conventional LiDAR odometry degrades in off-road and orchard conditions~\cite{Chen_2020_RAL,huaman2025comparative}. We re-implement DLO as a standalone (non-ROS) library to (i) keep the pipeline modular and (ii) inject a vehicle prior tailored to agricultural driving, which is dominated by wheel slip, uneven ground, and sharp local maneuvers.

Let $\mathbf{P}_k=\{\mathbf{p}^k_i\}_{i=1}^{N_k}$ be the $k$-th LiDAR sweep with pose $\mathbf{X}_k \in \mathrm{SE}(3)$. From past keyframes we maintain an active local submap $\mathcal{S}_{k-1}$, selected using nearest-neighbor and convex-hull criteria. The frontend estimates the relative motion $\hat{\Delta\mathbf{X}}_k$ between $\mathbf{P}_k$ and $\mathcal{S}_{k-1}$ using Generalized ICP (GICP).

Standard DLO assumes constant body velocity during a LiDAR sweep. We instead use a second-order motion prior that includes both velocity and acceleration of the platform. Let $\boldsymbol{\xi}(t)\in\mathfrak{se}(3)$ be the instantaneous body twist (linear and angular velocity) and $\dot{\boldsymbol{\xi}}(t)$ its time derivative (accelerations). The pose at time $t+\Delta t$ is predicted as
\begin{equation}
\hat{\mathbf{X}}(t+\Delta t)
\approx
\mathbf{X}(t)\,
\exp\!\Big(
\Delta t\,\boldsymbol{\xi}(t)
+ \tfrac{1}{2}\Delta t^2\,\dot{\boldsymbol{\xi}}(t)
\Big),
\label{eq:motion_model}
\end{equation}
where $\exp(\cdot)$ is the exponential map from $\mathfrak{se}(3)$ to $\mathrm{SE}(3)$. This model is used both as the registration prior and for LiDAR deskewing. In practice it stabilizes convergence when the tractor undergoes sudden acceleration or pitch/roll excursions on rough soil.

\subsubsection{SLAM Backend}
\label{sec:slam_backend}
The backend maintains a factor graph over keyframe poses and solves it incrementally with iSAM2. The first keyframe $\mathbf{X}_0$ is anchored by a prior factor with covariance $\boldsymbol{\Sigma}_{\mathrm{prior}}$. For each new keyframe $k$, we add an odometry factor that enforces consistency between $(\mathbf{X}_{k-1}, \mathbf{X}_k)$ and the frontend estimate $\hat{\Delta\mathbf{X}}_k$. The residual is computed in the Lie algebra of $\mathrm{SE}(3)$ via the log map, and is weighted by an odometry covariance $\boldsymbol{\Sigma}_{\mathrm{odo}}$.

Loop closures are introduced as non-consecutive pose constraints. Loop candidates are first proposed using (i) a KD-tree search in pose space and (ii) Scan Context descriptors~\cite{kim2018scan}. Candidates are ranked by cosine similarity between descriptors, then verified geometrically with GICP. Only matches that pass GICP are inserted as loop-closure factors. This reduces long-term drift and enforces global consistency of the trajectory.



\subsection{Multi-View 3D Gaussian Splatting}
\label{sec:gaussian_splatting}

We extend 3DGS to a synchronized multi-view setting to handle occlusions in orchards. The platform carries multiple cameras with complementary views (Fig.~\ref{fig:setup}). At each keyframe, all cameras are optimized jointly: their gradients update a \emph{single shared} set of splats. Gaussians are not duplicated per camera; each splat lives once in the global map and may be seen by one or more views.

Each Gaussian $i$ is defined by
\begin{equation}
\boldsymbol{\mu}_i \in \mathbb{R}^3,\quad
\boldsymbol{\Sigma}_i \in \mathbb{R}^{3 \times 3},\quad
o_i \in \mathbb{R},\quad
\mathbf{c}_i \in \mathbb{R}^{d_{\mathrm{SH}}},
\end{equation}
where $\boldsymbol{\mu}_i$ is its 3D center, $\boldsymbol{\Sigma}_i$ its 3D covariance, $o_i$ its opacity, and $\mathbf{c}_i$ spherical-harmonics (SH) color coefficients.

As in standard 3DGS, covariance comes from a learned scale $\mathbf{S}_i$ and rotation $\mathbf{R}_i$:
\begin{equation}
\boldsymbol{\Sigma}_i
= \mathbf{R}_i \, \mathbf{S}_i \, \mathbf{S}_i^{\top} \, \mathbf{R}_i^{\top}.
\end{equation}

For each camera $b \in \{1,\dots,B\}$, splats are projected into that camera via approximate rasterization. With world-to-camera transform $\mathbf{W}_b$ and projection Jacobian $\mathbf{J}_b$, the image-space covariance is
\begin{equation}
\boldsymbol{\Sigma}'_{i,b}
= \mathbf{J}_b \, \mathbf{W}_b \, \boldsymbol{\Sigma}_i \, \mathbf{W}_b^{\top} \, \mathbf{J}_b^{\top},
\end{equation}
and its $2 \times 2$ top-left block is used for rendering in view $b$.


Joint multi-view supervision lets all cameras refine the same map and the camera poses, improving consistency under dense foliage and repetitive row structure without per-camera splat copies.

\begin{table*}[t]
\vspace{1.5em}
\centering
\caption{Comparison of SLAM methods across three phenological stages for Apple Orchard and Pear Orchard.}\label{tab:qualitative_results}

\renewcommand{\arraystretch}{0.8}
\setlength{\tabcolsep}{0pt} 

\resizebox{\textwidth}{!}{%
\begin{tabular}{@{}ccccccccc@{}}
    \toprule
    & & {\scriptsize\textbf{Ground Truth}} & \textbf{AgriGS-SLAM} & \textbf{DLO+3DGS} & \textbf{OpenGS-SLAM} & \textbf{PINGS} & \textbf{Splat-SLAM} & \textbf{Photo-SLAM} \\
    \midrule
    {\scriptsize\multirow{3}{*}{\rotatebox{90}{\textbf{Apple Orchard}}}} &
      {\scriptsize\rotatebox{90}{\textit{Dormancy}}} &
      \includegraphics[width=0.12\textwidth]{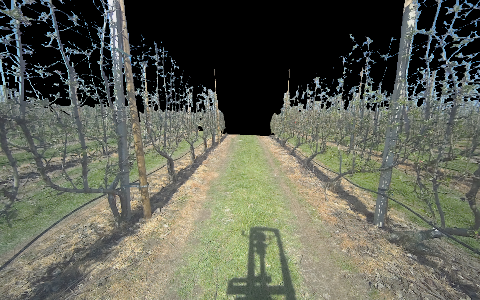} &
      \includegraphics[width=0.12\textwidth]{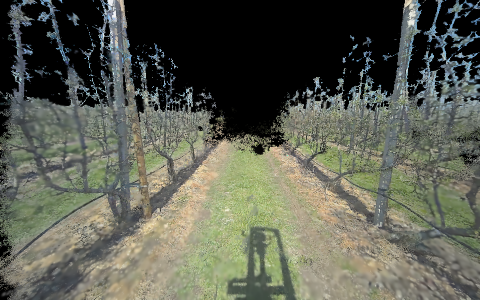} &
      \includegraphics[width=0.12\textwidth]{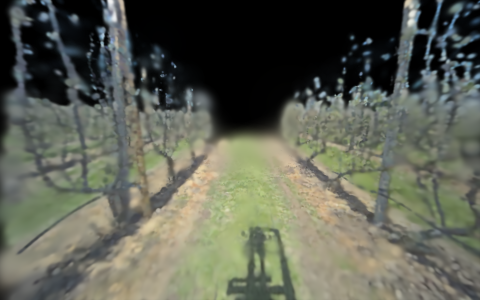} &
      \includegraphics[width=0.12\textwidth]{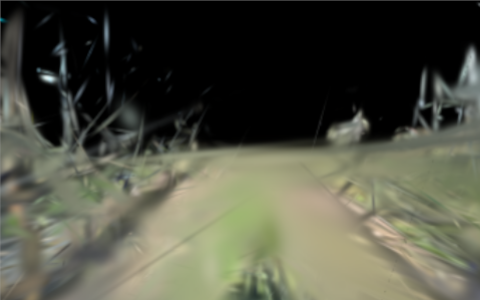} &
      \includegraphics[width=0.12\textwidth]{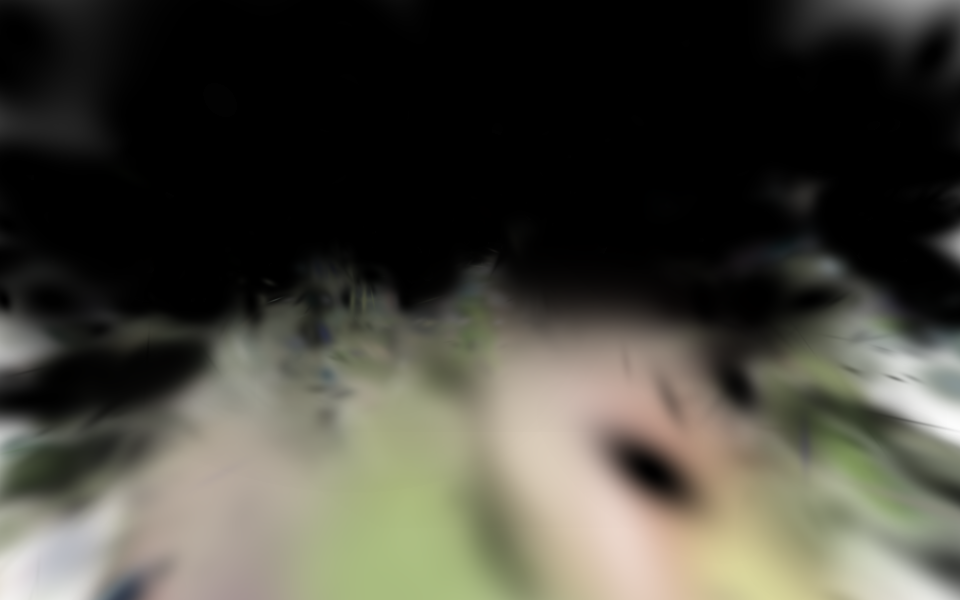} &
      \includegraphics[width=0.12\textwidth]{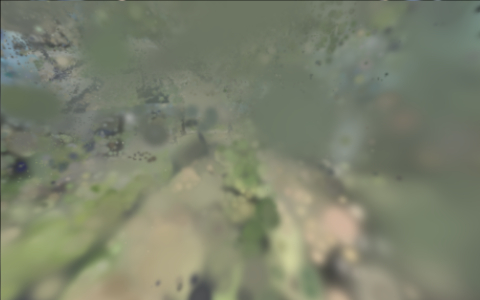} &
      \includegraphics[width=0.12\textwidth]{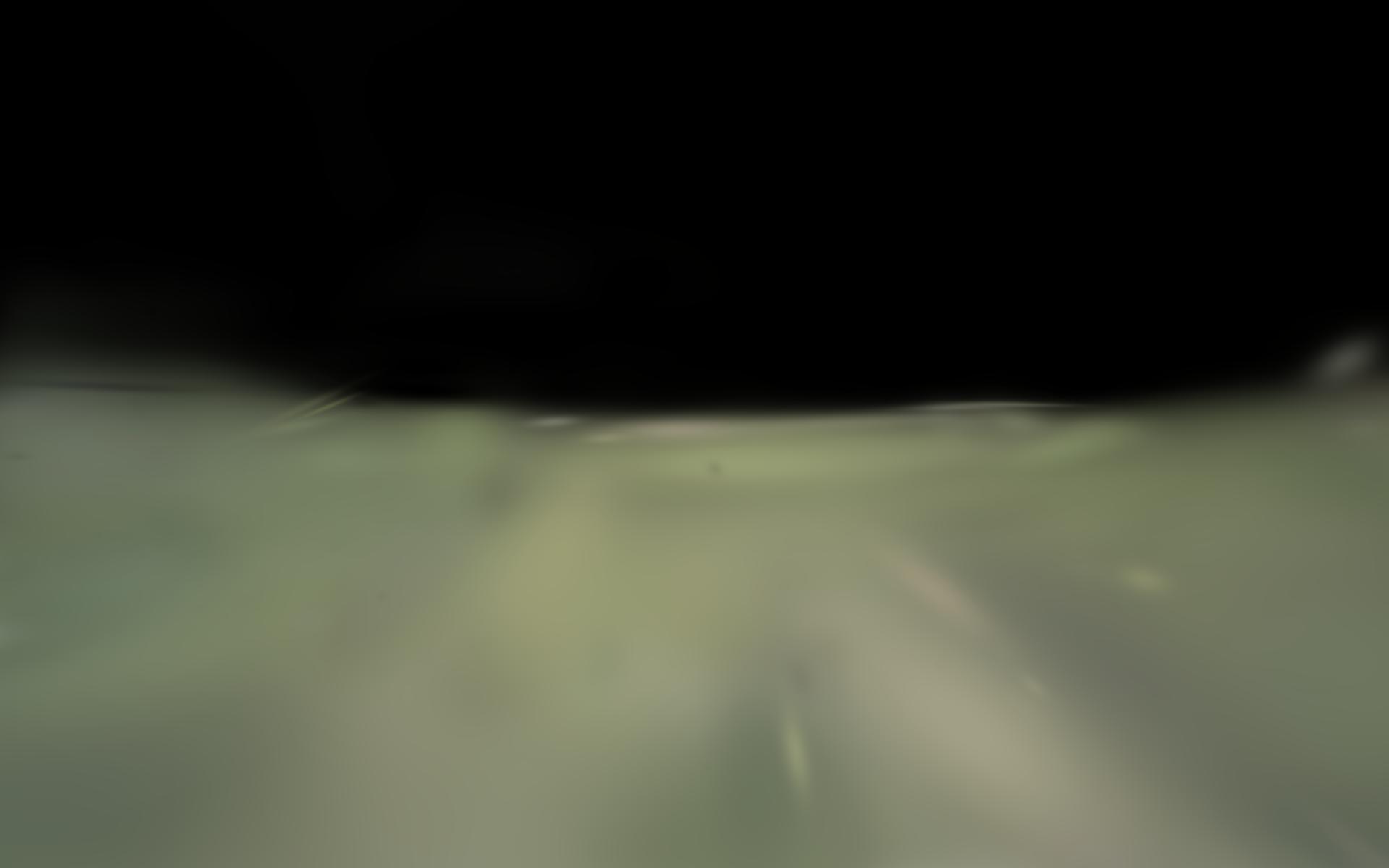}  \\
    & {\scriptsize\rotatebox{90}{\textit{Flowering}}} &
      \includegraphics[width=0.12\textwidth]{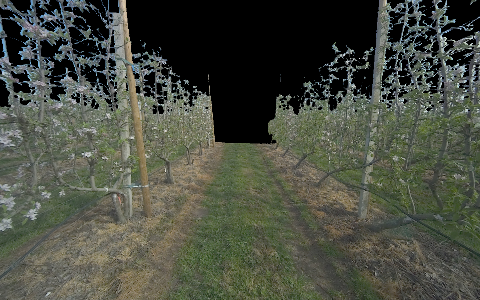} &
      \includegraphics[width=0.12\textwidth]{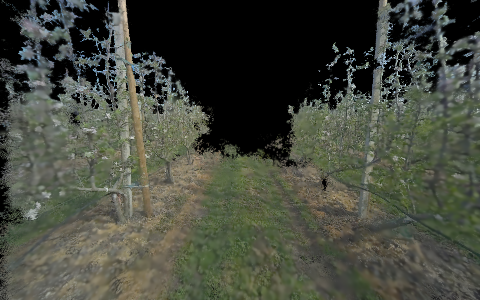} &
      \includegraphics[width=0.12\textwidth]{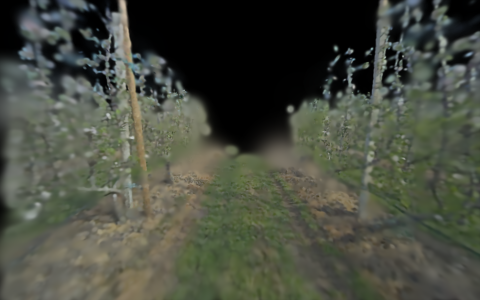} &
      \includegraphics[width=0.12\textwidth]{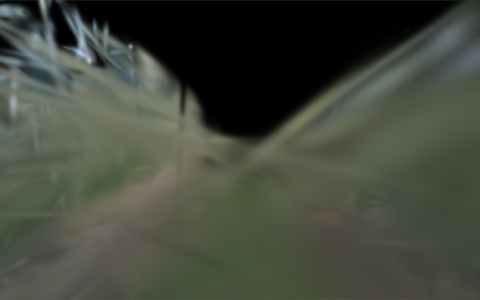} &
      \includegraphics[width=0.12\textwidth]{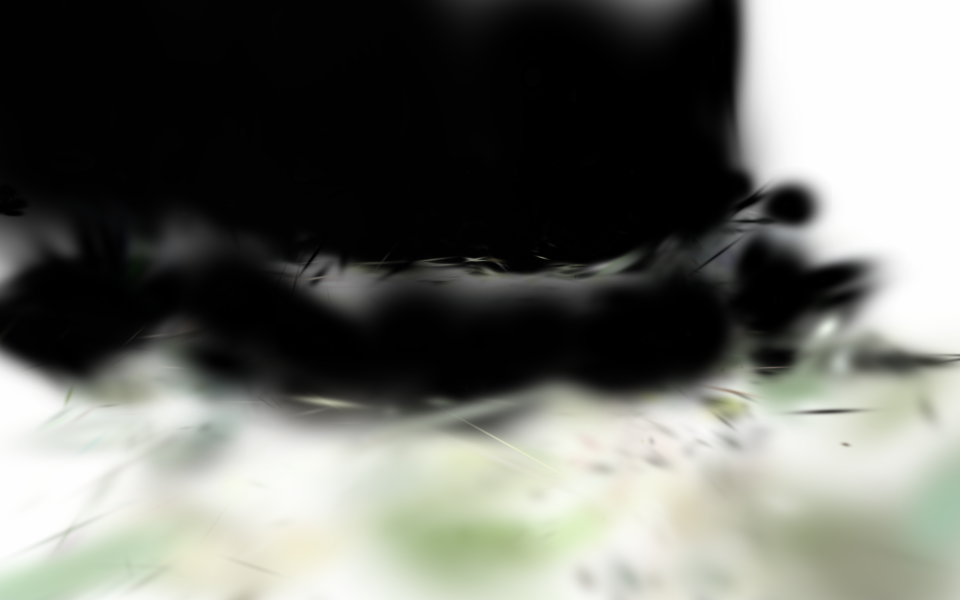} &
      \includegraphics[width=0.12\textwidth]{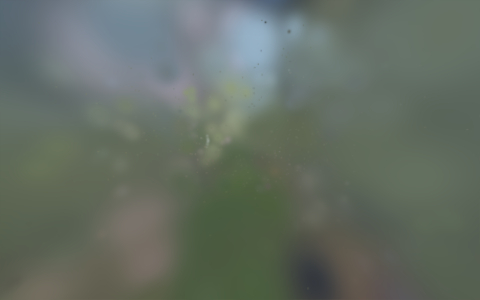} &
      \includegraphics[width=0.12\textwidth]{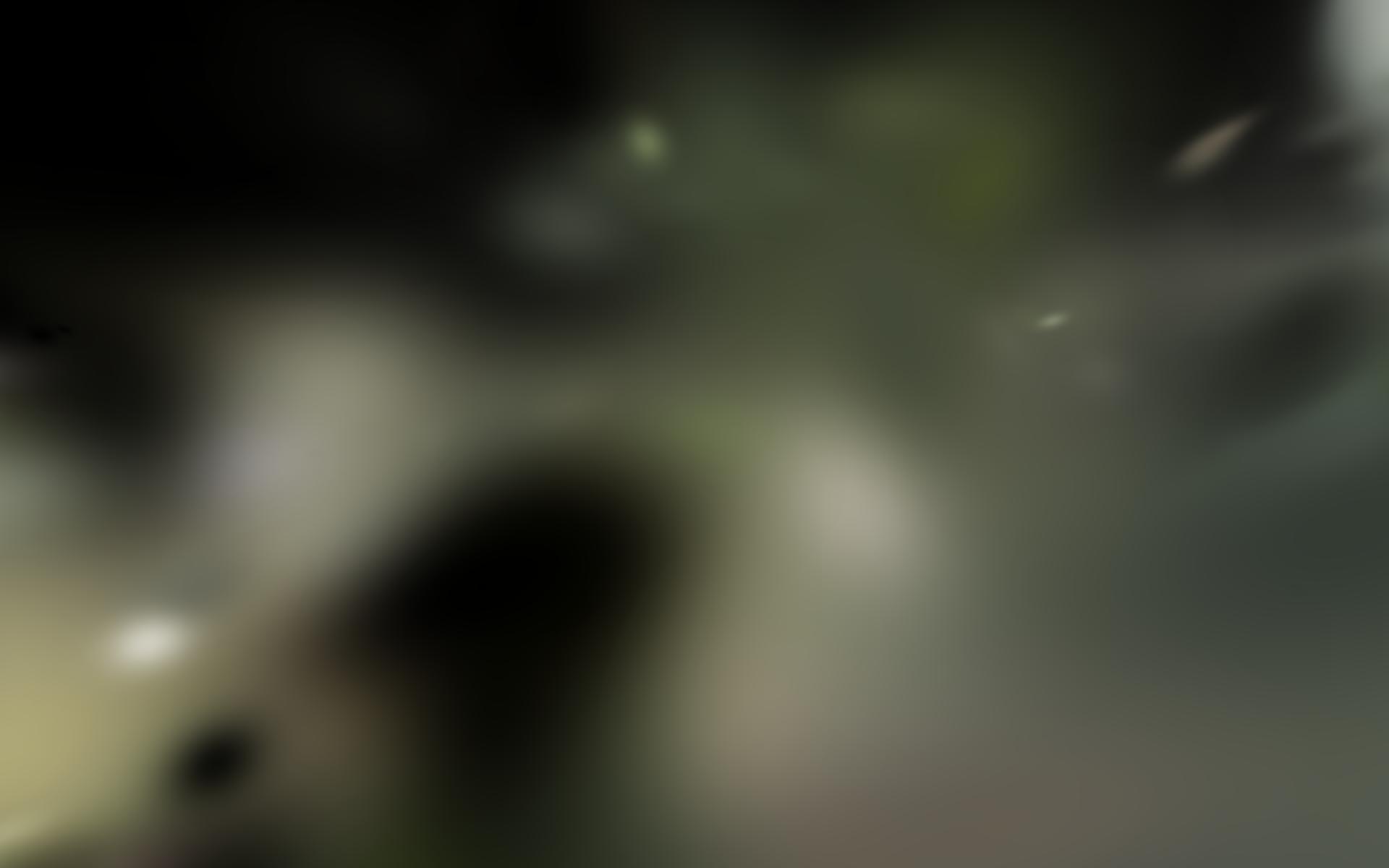}  \\
    & {\scriptsize\rotatebox{90}{\textit{Harvesting}}} &
      \includegraphics[width=0.12\textwidth]{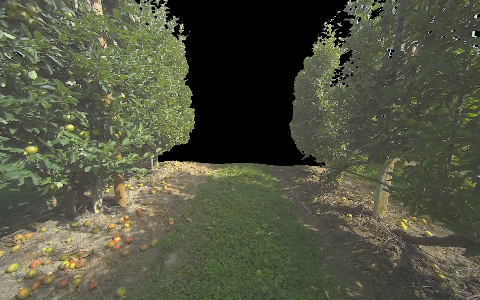} &
      \includegraphics[width=0.12\textwidth]{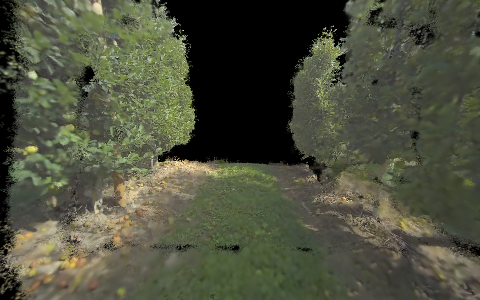} &
      \includegraphics[width=0.12\textwidth]{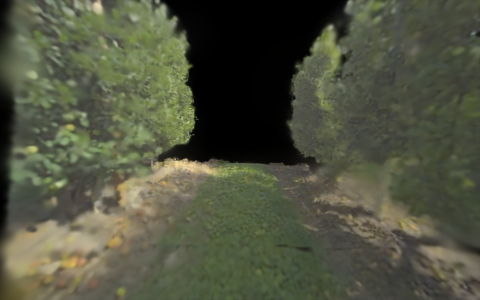} &
      \includegraphics[width=0.12\textwidth]{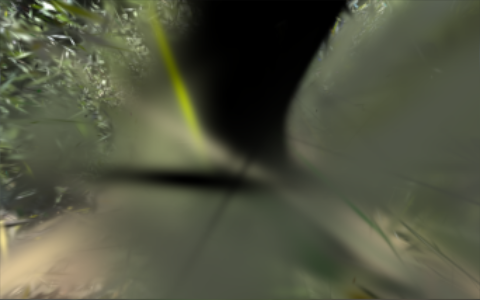} &
      \includegraphics[width=0.12\textwidth]{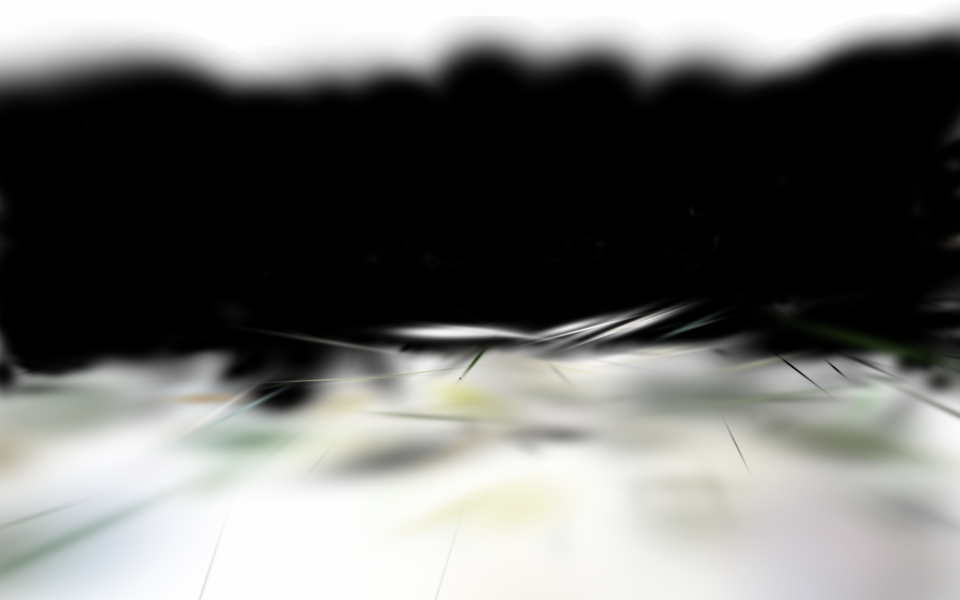} &
      \includegraphics[width=0.12\textwidth]{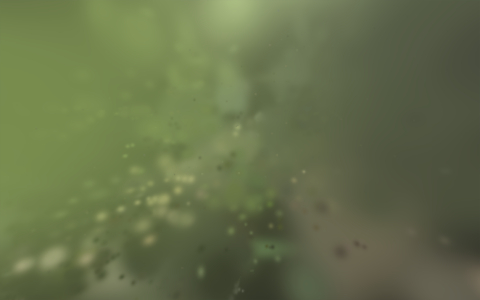} &
      \includegraphics[width=0.12\textwidth]{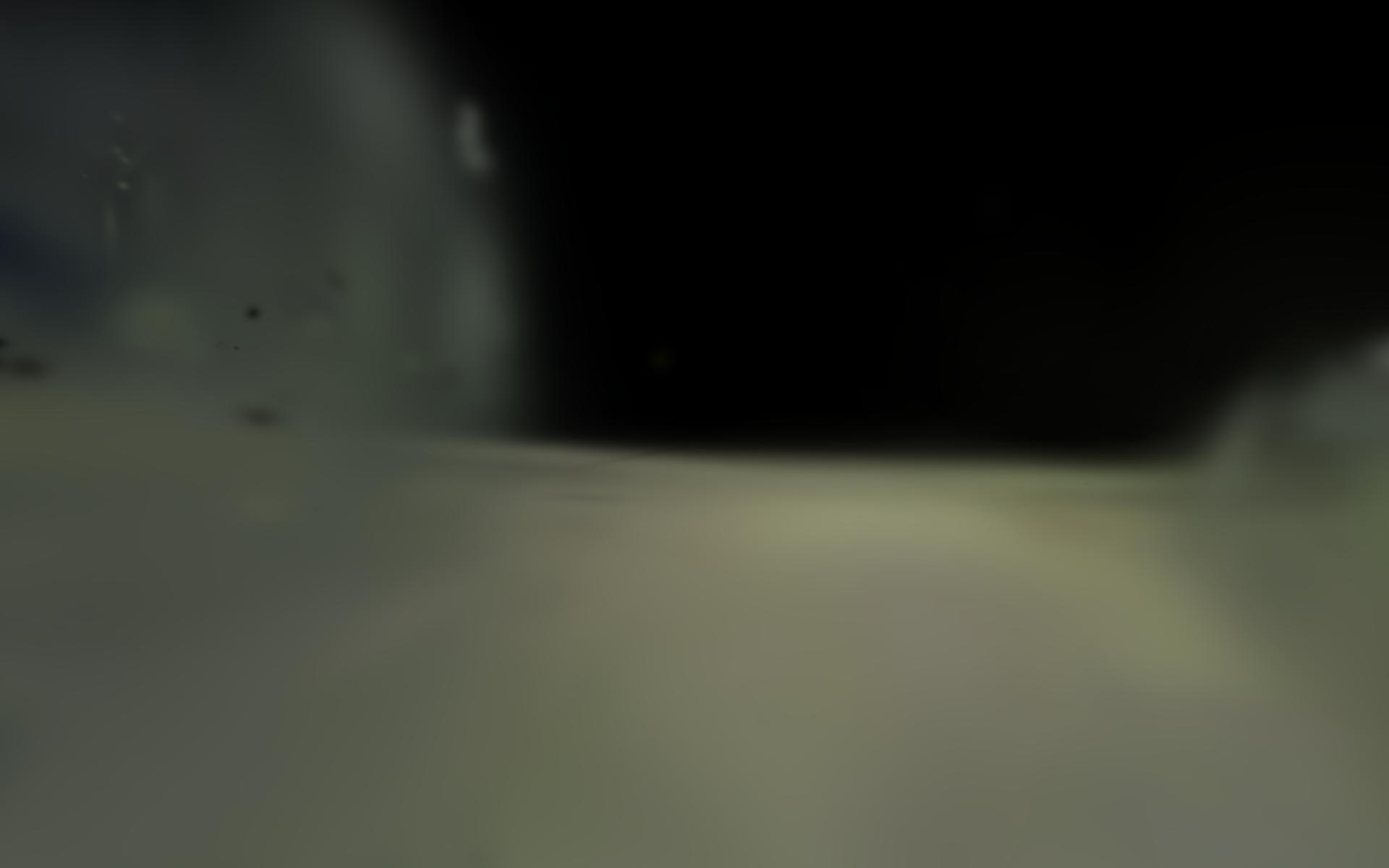}  \\
    \midrule
    {\scriptsize\multirow{3}{*}{\rotatebox{90}{\textbf{Pear Orchard}}}} &
      {\scriptsize\rotatebox{90}{\textit{Dormancy}}} &
      \includegraphics[width=0.12\textwidth]{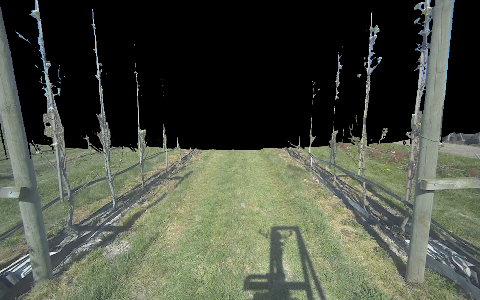} &
      \includegraphics[width=0.12\textwidth]{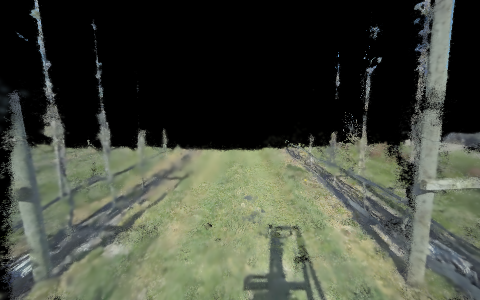} &
      \includegraphics[width=0.12\textwidth]{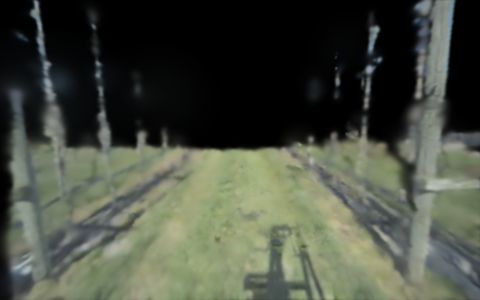} &
      \includegraphics[width=0.12\textwidth]{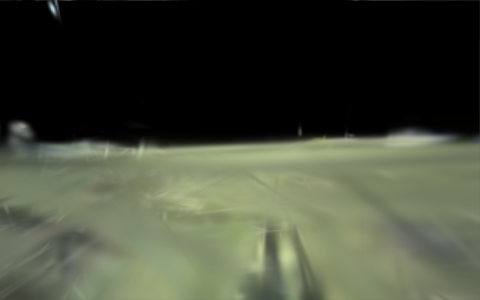} &
      \includegraphics[width=0.12\textwidth]{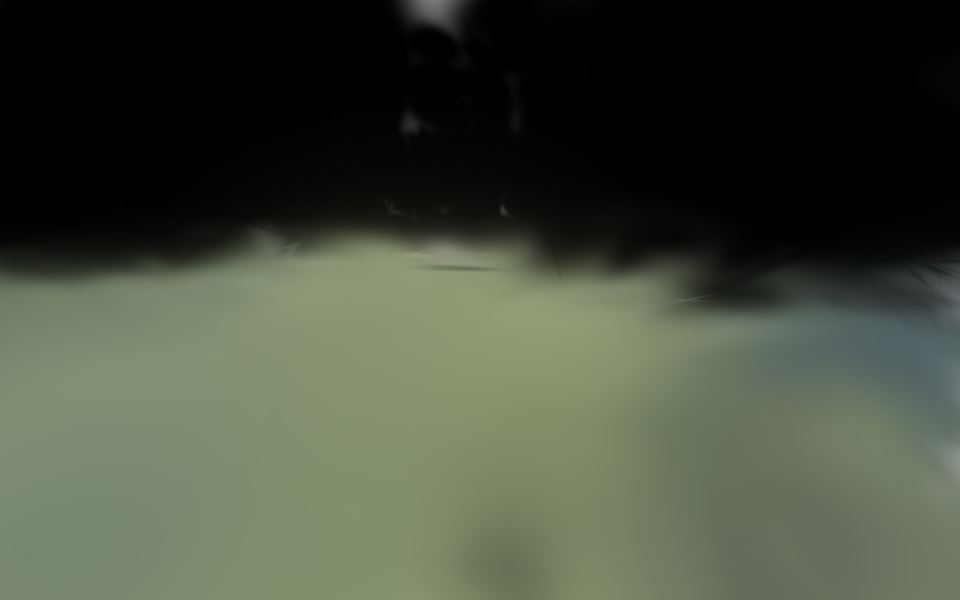} &
      \includegraphics[width=0.12\textwidth]{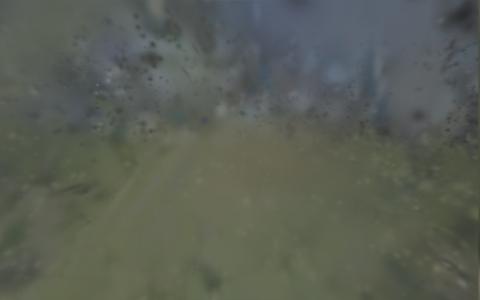} &
      \includegraphics[width=0.12\textwidth]{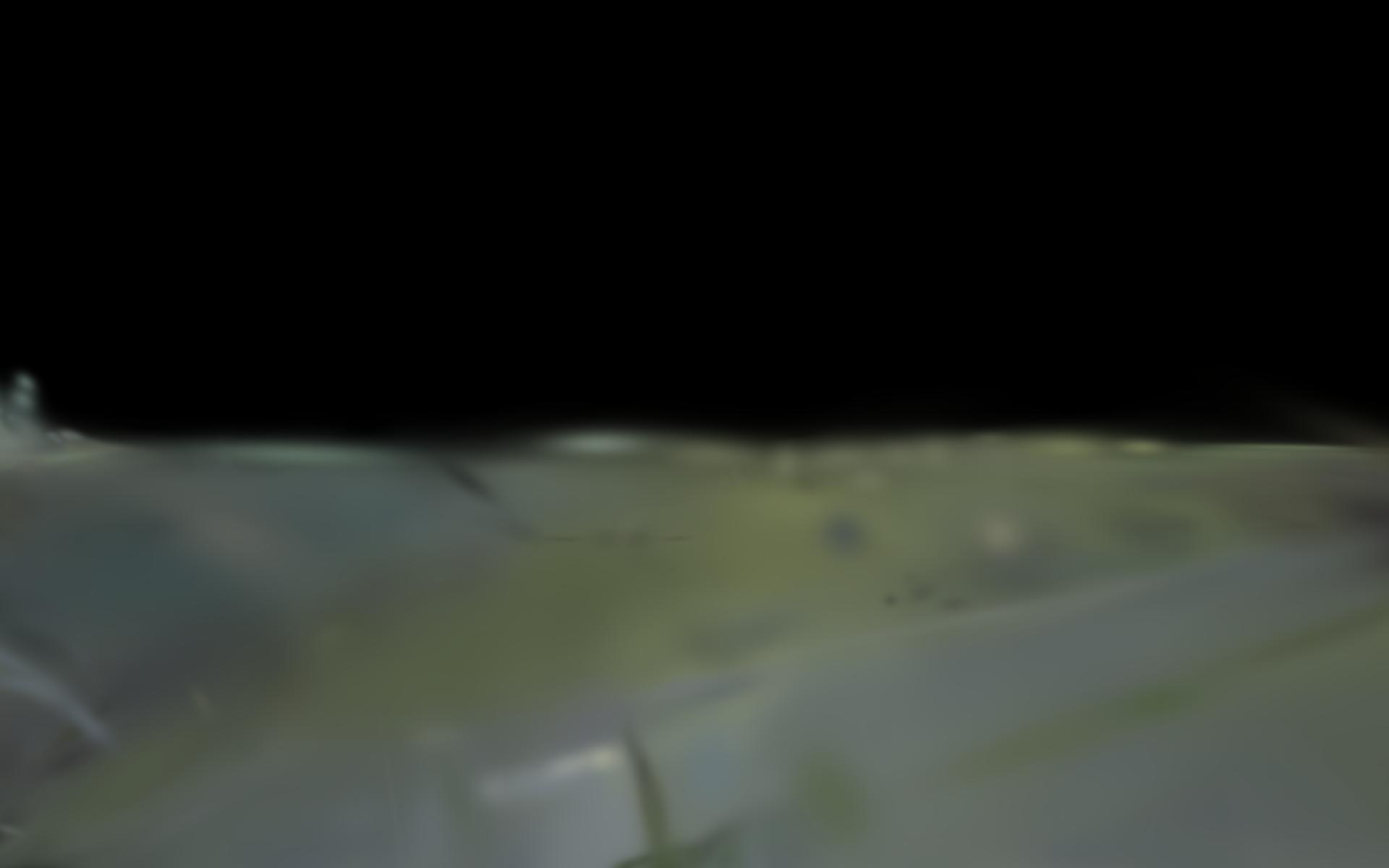}    \\
    & {\scriptsize\rotatebox{90}{\textit{Flowering}}} &
      \includegraphics[width=0.12\textwidth]{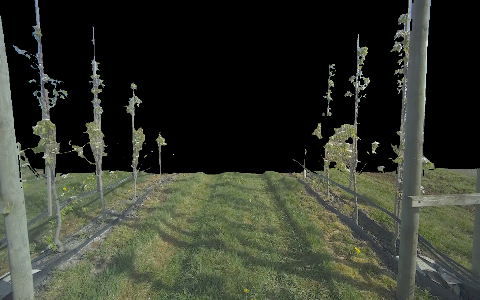} &
      \includegraphics[width=0.12\textwidth]{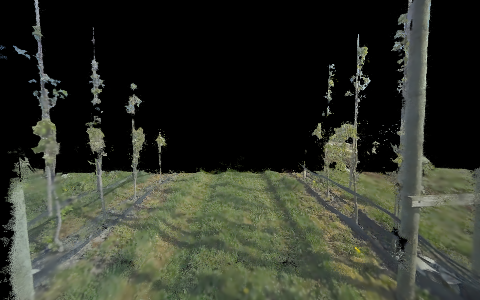} &
      \includegraphics[width=0.12\textwidth]{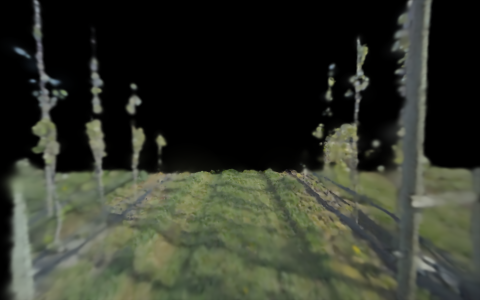} &
      \includegraphics[width=0.12\textwidth]{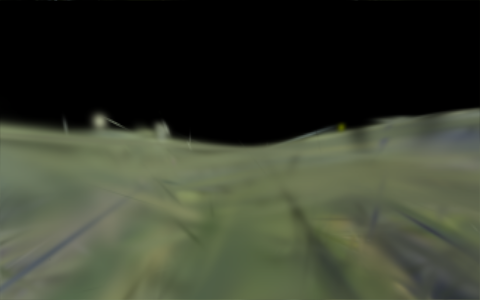} &
      \includegraphics[width=0.12\textwidth]{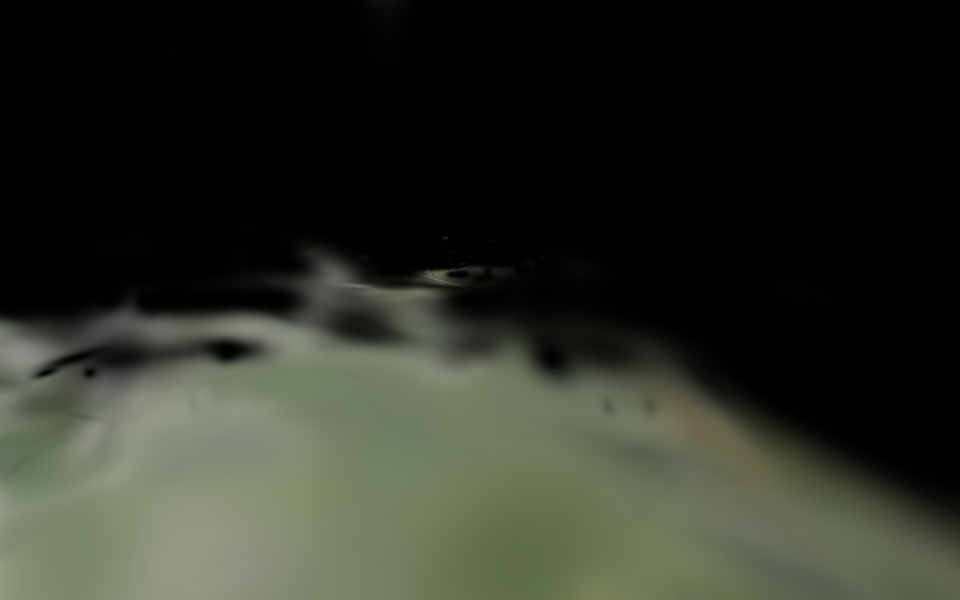} &
      \includegraphics[width=0.12\textwidth]{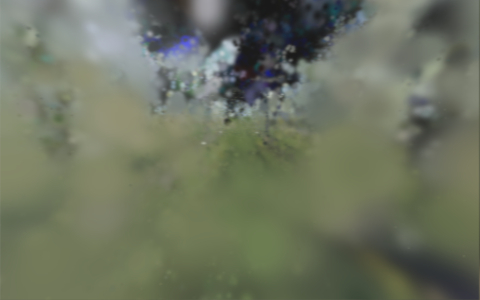} &
      \includegraphics[width=0.12\textwidth]{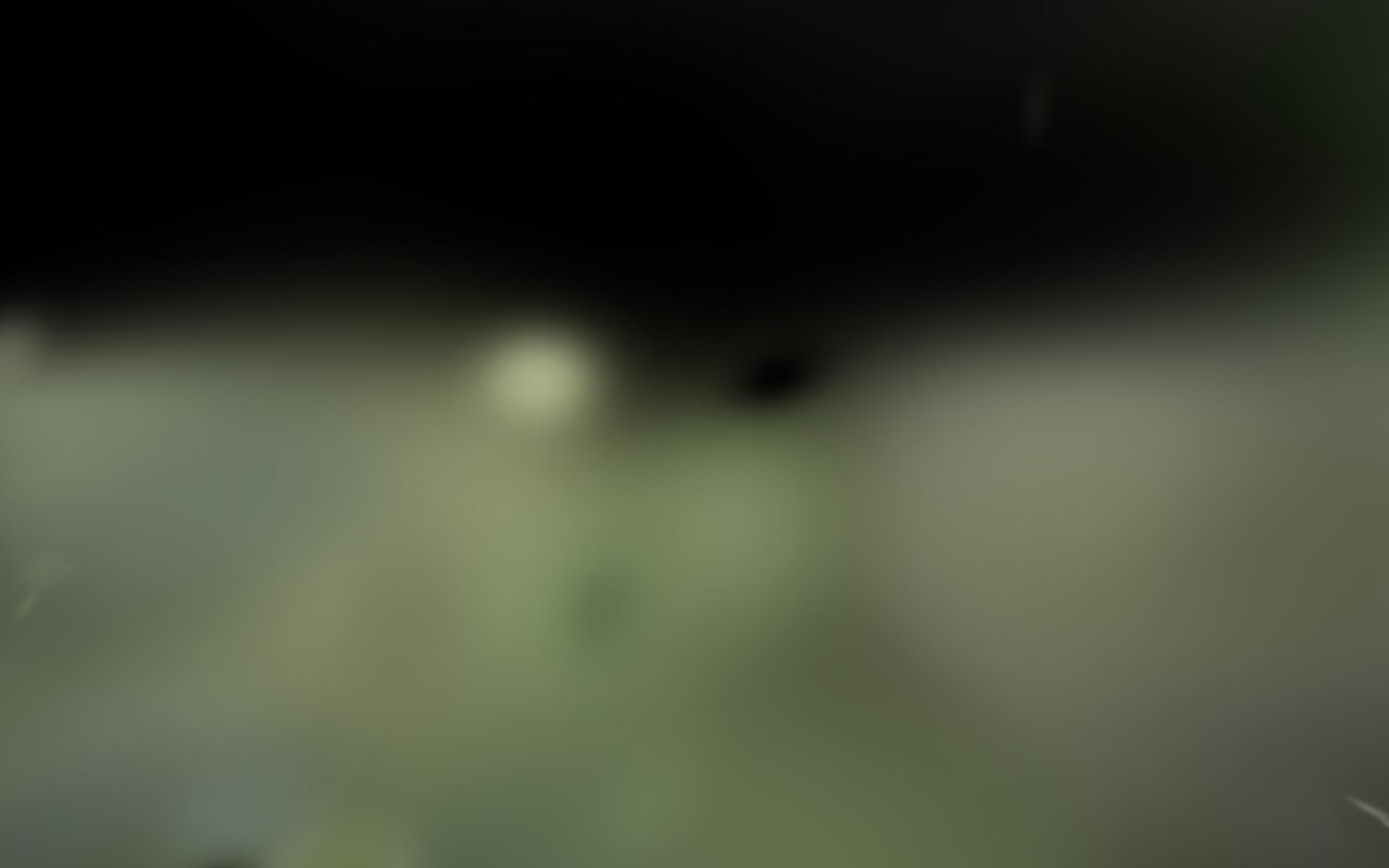}  \\
    & {\scriptsize\rotatebox{90}{\textit{Harvesting}}} &
      \includegraphics[width=0.12\textwidth]{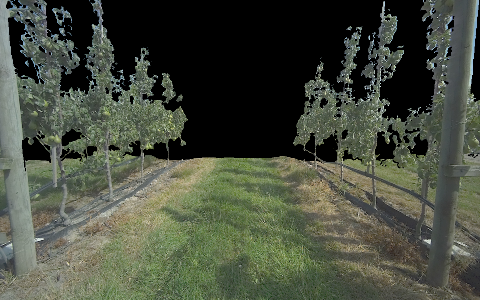} &
      \includegraphics[width=0.12\textwidth]{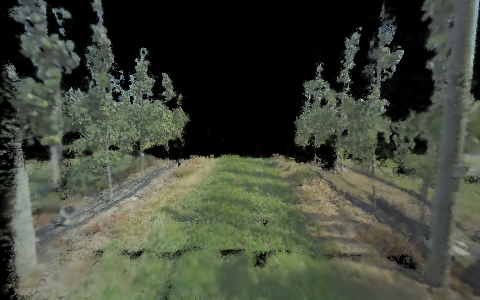} &
      \includegraphics[width=0.12\textwidth]{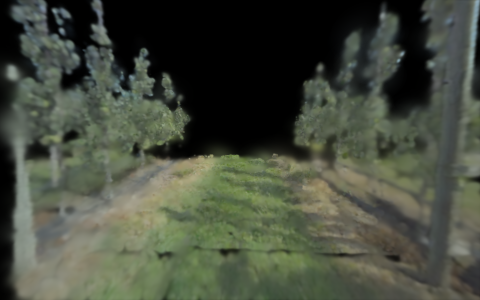} &
      \includegraphics[width=0.12\textwidth]{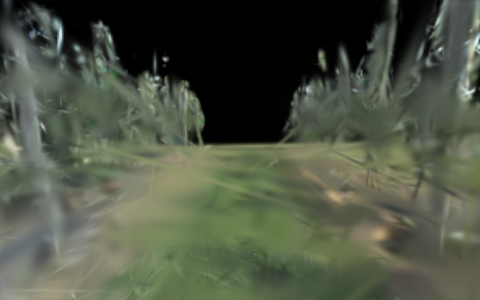} &
      \includegraphics[width=0.12\textwidth]{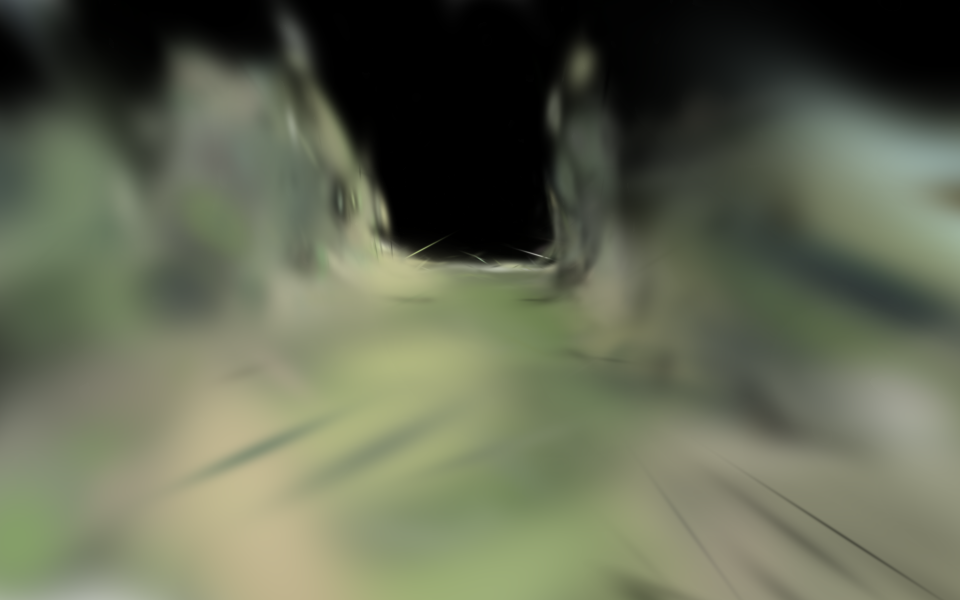} &
      \includegraphics[width=0.12\textwidth]{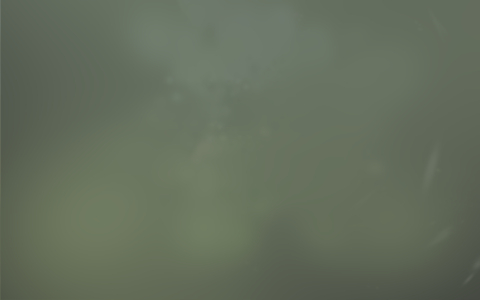} &
      \includegraphics[width=0.12\textwidth]{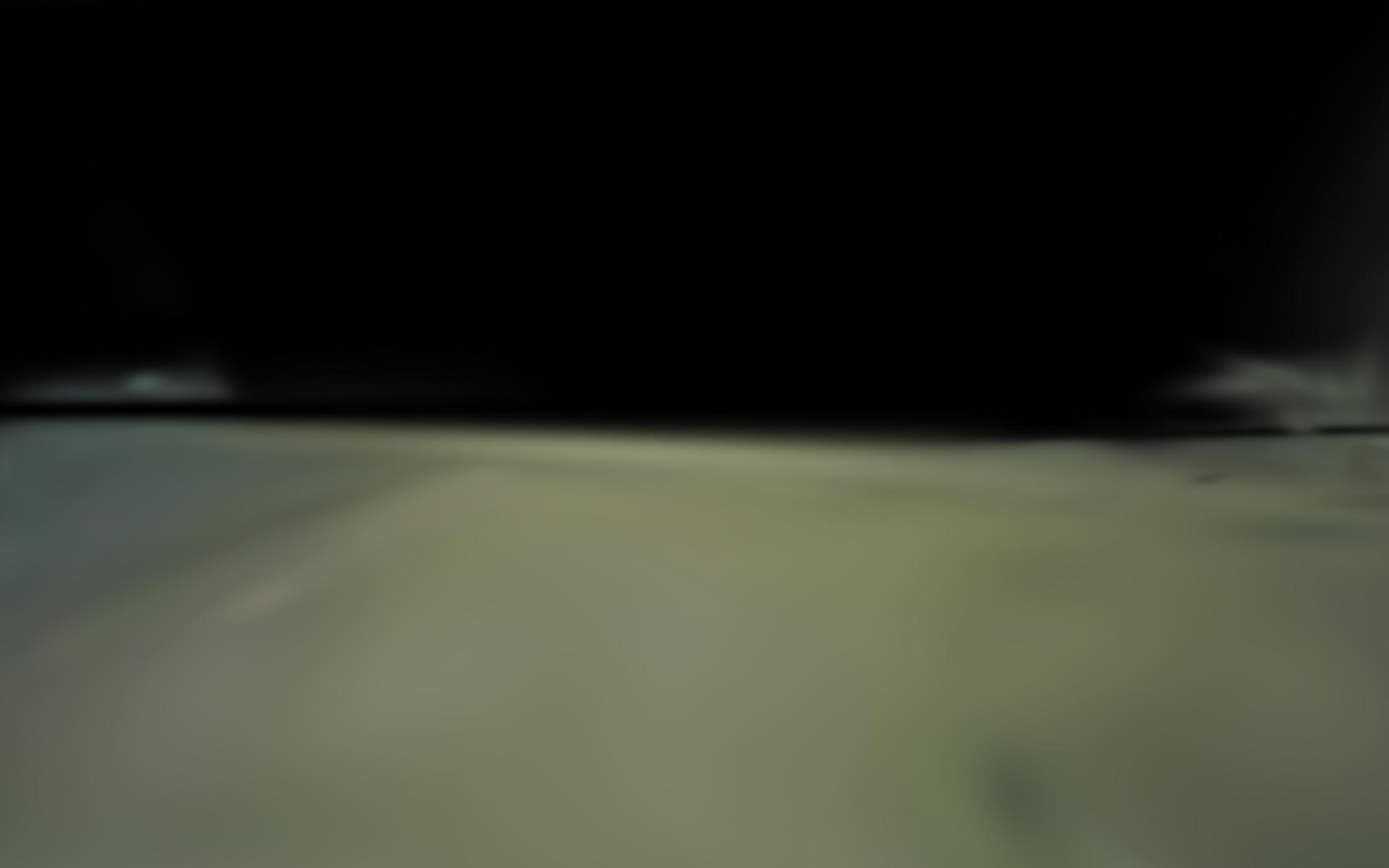}  \\
    \bottomrule
\end{tabular}%
} 

\renewcommand{\arraystretch}{1.0}
\setlength{\tabcolsep}{6pt} 
\vspace{-1.5em}
\end{table*}

\subsubsection{Incremental Gaussian Mapping}
\label{sec:incremental_gaussian_mapping}

The incremental Gaussian mapper maintains an explicit scene representation as a set of splats, stored in CPU memory for long--term retention and partially streamed to the GPU as an active set for real--time optimization.  

Whenever the SLAM system selects a new LiDAR keyframe, the corresponding RGB--D images from onboard cameras are refined to the LiDAR frame via GICP.  
This alignment yields a dense set of 3D points that (i) provide an initialization for the 3DGS process and (ii) compensate for small timestamp offsets across sensors, since camera and LiDAR frames recorded simultaneously on the same unit may exhibit slight delays.  

The fused points $\{\mathbf{p}_i\}$ are inserted into the submap by hashing each point into a voxel index:
\begin{equation}
k_i \;=\; \bigl\lfloor p_{i,x}/v \bigr\rfloor P_1
       \;+\; \bigl\lfloor p_{i,y}/v \bigr\rfloor P_2
       \;+\; \bigl\lfloor p_{i,z}/v \bigr\rfloor P_3,
\end{equation}
where $v$ is the current voxel size and $\{P_1,P_2,P_3\}$ are large primes chosen to minimize collisions.  
This voxel hashing strategy enables efficient coarse retrieval in large environments, where fast access to local neighborhoods is critical.  
Points whose voxel keys match existing splats are marked as active, while unmatched points instantiate new splats.  

\subsubsection{Memory Management}
To bound GPU memory usage, the mapper employs an adaptive active--set strategy.  
When the number of active splats exceeds a threshold $\tau$, the least--recently--used elements are offloaded to CPU memory.  
Conversely, splats stored on the CPU are reloaded when their voxel keys collide with new incoming points.  
This policy sustains real-time operation while retaining a complete global map, ensuring scalability along long orchard rows and supporting loop closures at revisited locations.

\subsubsection{Gaussian Splats Lifecycle Operations}  
The Gaussian map is updated online between keyframes. For each splat $j$, three lifecycle operations maintain detail while keeping memory bounded.  

\noindent \textbf{Pruning.}  
Every $\Delta_p$ iterations, the opacity is updated as the sigmoid of its trainable logit $\omega_j$. A splat is removed if its opacity $o_j$ is below $\tau_{\mathrm{opa}}=0.01$, or if its size $s_j$ is either too large ($s_j>0.3$) or too small ($s_j<5\times 10^{-5}$). This discards dim, unstable, or oversized splats while retaining potentially useful weak ones.  

\noindent \textbf{Densification.}  
Every $\Delta_d$ iterations, the image--space gradient magnitude at the projected splat center is computed. If the gradient exceeds a threshold $\gamma_g$, refinement is triggered: when the splat is relatively large ($s_j>\kappa$) it is split into smaller ones, whereas when the splat is still small ($s_j \leq 2s_{\mathrm{init}}$) it is duplicated. This increases local density around fine structures such as leaves and thin branches, without unbounded growth.  

\noindent \textbf{Opacity Reset.}  
Every $\Delta_r$ iterations, splats that are weak but not clear outliers are reset. Specifically, if the opacity lies between $\tau_{\mathrm{opa}}$ and a reset threshold $\tau_{\mathrm{reset}}$, and the size is within valid bounds ($s_{\min} < s_j < s_{\max}$), the opacity is refreshed. This allows low-opacity splats to recover under changing illumination and motion.  

\subsubsection{Optimization Loop}
At each iteration, the mapper schedules: opacity reset every $\Delta_r$ steps, pruning every $\Delta_p$ steps, densification every $\Delta_d$ steps, and GPU memory cleanup every $10$ steps.  
These routines are designed to run in real time between keyframes, combining gradient-driven densification, conservative pruning, and periodic resets to preserve fine detail while keeping memory usage bounded.

\subsection{Multimodal Loss Function}
\label{sec:multi_modal_loss}

We propose a composite objective that jointly aligns the rendered Gaussian splats with ground‐truth RGB imagery and LiDAR observations.  
Let \(I_{\mathrm{pred}}\) and \(I_{\mathrm{gt}}\) denote the rendered and ground‐truth RGB images, respectively;

\subsubsection{Photometric Consistency}
Following Kerbl et al.~\cite{kerbl20233d}, the parameters of the 3D Gaussians are optimized by minimizing a weighted combination of an \(L_1\) loss and a Structural Similarity Index (SSIM) loss between rendered and ground‐truth images. The balance between these two terms is controlled by \(\lambda\):
\begin{equation}
L_{\mathrm{3DGS}} =
(1-\lambda)\,\|I_{\mathrm{pred}} - I_{\mathrm{gt}}\|_{1}
+ \lambda\bigl(1 - \mathrm{SSIM}(I_{\mathrm{pred}}, I_{\mathrm{gt}})\bigr).
\end{equation}

\subsubsection{Probabilistic Depth Consistency}
To handle LiDAR sparsity and noise, we introduce a probabilistic depth-consistency loss.  
Each LiDAR return at pixel $u$ is modeled as a Gaussian with mean $D_{\mathrm{LiDAR}}(u)$ and fixed variance $\sigma_L^2$, while the renderer predicts a Gaussian $\mathcal{N}(\mu_u,\sigma_u^2)$.  
The loss averages the KL divergence between these distributions over all valid pixels $\Omega$:

\begin{equation}
L_{\mathrm{LiDAR}}
=
\lambda_{L}\,\frac{1}{|\Omega|}
\sum_{u \in \Omega}
\mathrm{KL}\!\big(
\mathcal{N}(D_{\mathrm{LiDAR}}(u), \sigma_L^2)
\;\|\;
\mathcal{N}(\mu_u, \sigma_u^2)
\big).
\end{equation}

This enforces alignment between LiDAR geometry and rendered depth, enabling sensor fusion at the loss level.

\subsubsection{Total Multimodal Loss}
The final objective $L_{\mathrm{AgriGS}}$ combines photometric and geometric supervision:
\begin{equation}
L_{\mathrm{AgriGS}} = L_{\mathrm{3DGS}} + L_{\mathrm{LiDAR}}.
\end{equation}

During training, the multimodal loss guides the simultaneous optimization of the Gaussian scene representation and the camera or robot poses, integrating complementary photometric and geometric cues for consistent supervision.


\section{Experiments}
\label{sec:experiments}

We conducted experiments across three phenological phases, dormancy, flowering, and harvesting, capturing the impact of seasonal changes in foliage, occlusions, and illumination. Two contrasting orchards were considered: an apple orchard with espalier canopies that simplify navigation, yet grow dense with leaves, and a pear orchard with globular trees whose volumetric structure creates persistent occlusions and spatial complexity. Data collection followed a standardized protocol: one clockwise pass for training and two counter-clockwise passes, the final reserved for novel-view validation (Fig.~\ref{fig:cover}). This design introduced viewpoint variation while maintaining temporal consistency and maximizing overlap across the multi-camera platform (Fig.~\ref{fig:setup}). To focus on orchard reconstruction, we explicitly remove the sky from observations, avoiding large uninformative regions that would bias optimization and degrade map consistency.

The literature on 3DGS-SLAM has expanded rapidly in the past year. To capture the breadth of current approaches, we selected four highly recent methods, published within the last 12 months at leading robotics and computer vision venues, balancing indoor and outdoor settings as well as diverse tracking strategies. For indoor environments, we included Photo-SLAM~\cite{Huang_2024_CVPR}, which extends the classical ORB-SLAM3~\cite{campos2021orb} pipeline with handcrafted feature tracking, and Splat-SLAM~\cite{sandstrom2024splatslam}, which leverages the deep learning-based tracking of DROID-SLAM~\cite{teed2021droid} to represent the neural counterpart of the classical paradigm. For outdoor settings, we selected OpenGS-SLAM~\cite{yu2025rgb}, which employs transformer-based depth prediction followed by RANSAC and PnP alignment for robust pose estimation, and PINGS~\cite{pan2025pings}, a LiDAR-based system that integrates scan-matching with SDF representations. We excluded categories beyond our scope, such as inertial-based methods requiring ROS/ROS 2 integration, as well as earlier works superseded by more recent contributions. Incremental variants (e.g., DROID-Splat~\cite{homeyer2024droid}) that offer only minor deviations or are outperformed were also omitted. Similarly, we excluded methods without public implementations (e.g., LSG-SLAM~\cite{xin2025large}) and those reliant on ground-truth supervision (e.g., Point-SLAM~\cite{sandstrom2023point}), as they are unsuitable for real-world deployment.

\begin{figure}[t]
\vspace{1.0em}
\centering
\includegraphics[width=0.47\textwidth]{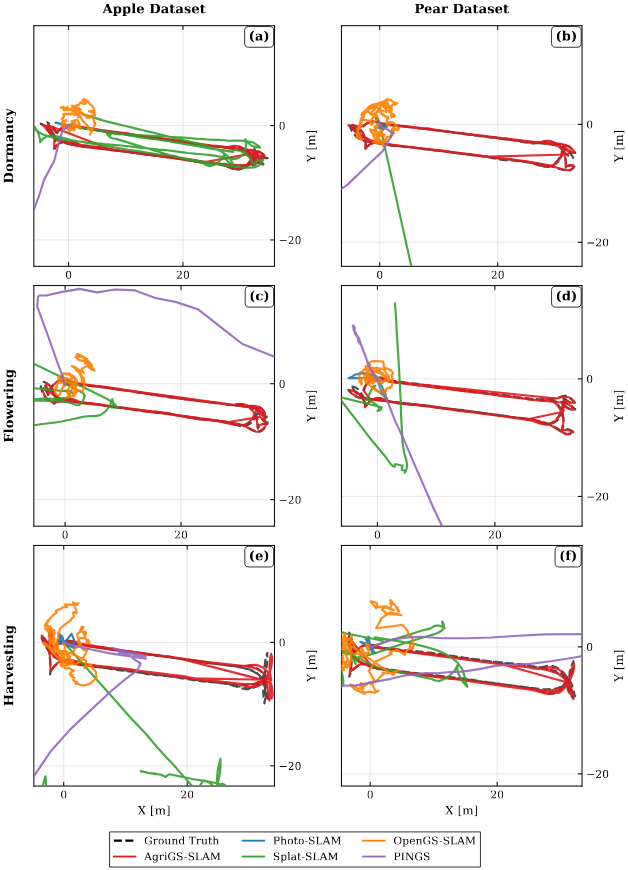}
\caption{Comparison of SLAM trajectories against ground truth.}\label{fig:traj}
\vspace{-2.0em}
\end{figure}

\section{Results and Discussion}
\label{sec:results}

Our primary baseline, DLO+3DGS, serves a twofold purpose.  
First, by disabling pose correction from 3DGS back to odometry, it isolates the effect of LiDAR-guided localization via $L_{\mathrm{LiDAR}}$, allowing a direct comparison between gradient-driven pose refinement and the DLO prior (see Subsection~\ref{subsec:ablation}~\S\ref{subsec:ablation:localization}).  
Second, it benchmarks the 3DGS training strategy introduced in Section~\ref{sec:proposed_approach} (see Subsection~\ref{subsec:ablation}~\S\ref{subsec:ablation:mapping}) against the original schedule of Kerbl~et~al.~\cite{kerbl20233d}.

\subsection{Quantitative Results}
\label{sec:quantitative_results}
Overall, AgriGS-SLAM outperforms the state-of-the-art across both orchard scenarios (Table~\ref{tab:quantitative}). Performance is weaker in the dormancy domain, where the sparsity of trunks and thin branches makes optimization harder and produces more splatting artifacts. The highest results are obtained in the harvesting stage, with similarly strong results in flowering due to blossom density.

Multi-camera setups generally enhance localization accuracy (ATE), with two cameras outperforming one and three outperforming two. The sole exception arises in apple harvesting, where the DLO+3DGS baseline marginally surpasses AgriGS-SLAM despite the overall trend. Notably, although the three-camera configuration often achieves the best scores, the single-camera AgriGS-SLAM attains higher average performance across all metrics, underscoring the robustness of the baseline. This can be attributed to the fact that multiple overlapping views increase the likelihood of rendering artifacts, whereas the single-view setting avoids such inconsistencies while still delivering strong localization.

On novel-view trajectories, OpenGS-SLAM achieves the best results among state-of-the-art methods, with superior SSIM in pear flowering and apple dormancy. Nevertheless, its performance is consistently below AgriGS-SLAM in other novel-view and training-view metrics. Conversely, on training-view trajectories, Splat-SLAM is the strongest competitor, surpassing OpenGS-SLAM in both localization and rendering, though still inferior to AgriGS-SLAM. Importantly, Splat-SLAM diverged in one sequence (pear dormancy) even after extensive tuning, whereas AgriGS-SLAM completed all runs without failure.

Among the remaining baselines, DLO+3DGS obtains higher PSNR in novel view but performs poorly in SSIM and LPIPS due to noisier splats designed for long offline optimizations rather than our real-time 3DGS scheduling. PINGS achieves the worst results overall, despite being LiDAR-based and designed for outdoor environments. Finally, Photo-SLAM (ORB-SLAM3 based) proved unsuitable for orchards, consistently crashing even after extensive parameter tuning.

\subsection{Qualitative Results}
\label{subsec:qualitative}

Table~\ref{tab:qualitative_results} reports a visual comparison across different SLAM baselines in orchard environments under varying seasonal and structural conditions. Reconstructions are shown for both apple and pear orchards during dormancy, flowering, and harvesting.  

Across all conditions, Photo-SLAM and Splat-SLAM struggle to preserve structural consistency, producing overly blurred or distorted representations that fail to capture the tree rows and inter-row spacing. Similarly, PINGS frequently collapses the orchard geometry, resulting in severe artifacts and large black voids where structural details are lost. OpenGS-SLAM yields more coherent layouts, yet reconstructions remain unstable, with wavy ground surfaces and over-smoothed vegetation.  

In Fig.~\ref{fig:traj}, we show the 2D projection of the trajectories reported in Table~\ref{tab:quantitative}. Most methods fail in both orchard scenarios, with only Splat-SLAM achieving comparatively better results. Although proposed for outdoor tasks, PINGS yields the poorest performance, even underperforming the indoor-oriented methods Photo-SLAM and Splat-SLAM. Conversely, OpenGS-SLAM achieves the best rendering quality but fails to maintain accurate localization, despite being designed for outdoor environments.


\subsection{Ablation Studies}\label{subsec:ablation}
We perform ablation studies to assess the contribution of each component of our pipeline (\S\ref{subsec:ablation:localization}, ~\S\ref{subsec:ablation:mapping}, ~\S\ref{subsec:ablation:multicamera}); as part of this analysis, we also evaluated the effect of adding supplementary loss functions to the core objective (\S\ref{subsec:ablation:multimodalloss}), which did not provide further improvements and confirms the robustness of our design:

\begin{enumerate}
    \item \textbf{$\nabla$ Localization (Table~\ref{tab:quantitative}).}  \label{subsec:ablation:localization}
    A direct comparison to our primary baseline, DLO+3DGS, highlights the impact of the LiDAR supervision in AgriGS-SLAM. While the former achieves strong reconstructions, the latter improves both mapping and localization simultaneously. The effect is most pronounced in the dormancy and flowering phase, where $L_{\mathrm{LiDAR}}$ reduces ATE despite sparsity. Similar gains are observed during harvesting, where dense canopies often degrade odometry but AgriGS-SLAM preserves scene coherence and trajectory accuracy.
    
    \item \textbf{$\nabla$ Mapping (Table~\ref{tab:quantitative}).}  
    \label{subsec:ablation:mapping}
    We compare our proposed training schedule with the original formulation~\cite{kerbl20233d}. While keeping data, poses, and loss terms unchanged, our strategy consistently delivers improved rendering metrics across both seasons and orchard types, reflecting more stable optimization and a stronger coupling between geometry and appearance. 

    \item \textbf{Multi-Camera Views (Table~\ref{tab:slam_multiview_grouped_ocm}, Table~\ref{tab:quantitative})}
    \label{subsec:ablation:multicamera}
    The multi-camera ablation reveals a clear trend: while adding views progressively sharpens reconstructed geometry and reduces drift on training trajectories, it simultaneously degrades novel-view synthesis by increasing overfitting. 

    \item \textbf{Multimodal Loss (Table~\ref{tab:agrigs_ablation}).} 
    \label{subsec:ablation:multimodalloss}
    A complementary study on the multimodal loss evaluates the added contributions of Chamfer distance for point-cloud alignment, LOS constraints for geometry, surface-normal regularization for structure, and Raydrops for occlusion, with Table~\ref{tab:agrigs_ablation} reporting the average results of each sequence. The complete formulation consistently improves PSNR, SSIM, and LPIPS over reduced variants, while normal regularization proves ineffective in orchards with irregular surfaces and Chamfer distance adds overhead without clear benefits. 
\end{enumerate}

\begin{table}[t]
\vspace{1.0em}
\centering
\caption{Supplementary analysis of Multimodal Loss.}
\renewcommand{\arraystretch}{1.1}
\vspace{-1.0em}
\begin{tabular}{@{}lcccc@{}}
\toprule
\textbf{Method} & \textbf{PSNR}~$\uparrow$ & \textbf{SSIM}~$\uparrow$ & \textbf{LPIPS}~$\downarrow$ \\
\midrule
\textbf{AgriGS-SLAM}   & \textbf{26.7204} & \textbf{0.8870} & \textbf{0.2177} \\
AgriGS-SLAM w/ Chamfer      & 25.9230 & 0.8626 & 0.2638 \\
AgriGS-SLAM w/ LOS          & \underline{26.5652} & \underline{0.8827} & \underline{0.2290} \\
AgriGS-SLAM w/ Normals      & 25.8807 & 0.8544 & 0.2800 \\
AgriGS-SLAM w/ Raydrops     & 26.5247 & 0.8813 & 0.2292 \\
\bottomrule
\end{tabular}
\label{tab:agrigs_ablation}
\vspace{-2.0em}
\end{table}

\vspace{-0.5em}
\section{Conclusion}
\label{sec:conclusion}

We presented AgriGS-SLAM, a Visual--LiDAR 3DGS-SLAM framework for orchards that integrates direct LiDAR odometry, loop-closure refinement, and incremental memory-aware mapping under a multimodal optimization scheme. The multi-camera design mitigates occlusions, while a gradient-driven map lifecycle and a LiDAR-guided depth term couple geometry and appearance more effectively than prior pipelines. An orchard-specific trajectory template that evaluates both training- and novel-view synthesis reduces reconstruction bias. Tested across apple and pear orchards at multiple growth stages, AgriGS-SLAM achieves higher reconstruction quality and trajectory accuracy than recent 3DGS-SLAM baselines, with ablations confirming the impact of the LiDAR loss, scheduled training, and multi-camera input, while working real-time under seasonal variation. 

Limitations include sensitivity to LiDAR--camera calibration and reduced performance in sparser scenes, where misalignments may occur from sensor delays and motion. The KL-divergence depth-consistency term alleviates but cannot fully remove these cross-modal temporal discrepancies.

As future work, we plan to extend recordings to additional growth stages with semantic annotations (e.g., flowering and fruit counts), enabling broader use in AgriGS-SLAM, and to incorporate inertial cues for tighter motion priors and full on-tractor deployment. While compatible with digital-twin workflows, this study establishes improved SLAM and 3D reconstruction in real orchards, providing a foundation for precision-agriculture applications from on-tractor perception to farm-management support.

\addtolength{\textheight}{-12cm}   




\section*{Acknowledgment}
The authors thank the Fruit Research Center (FRC) in Randwijk for access to the orchards. Mirko Usuelli’s work was carried out within the Agritech National Research Center and funded by the EU Next-GenerationEU (PNRR – M4C2, Inv. 1.4 – D.D. 1032 17/06/2022, CN00000022). This manuscript reflects only the authors’ views; the EU and Commission are not responsible. Contributions from Matteo Matteucci, Gert Kootstra, and David Rapado-Rincon were co-funded by the EU Digital Europe Programme (AgrifoodTEF, GA Nº 101100622).

\bibliographystyle{IEEEtran}
\bibliography{IEEEexample}

\end{document}